\documentclass[conference]{IEEEtran}
\IEEEoverridecommandlockouts

\usepackage{cite}
\usepackage{amsmath,amssymb,amsfonts}
\usepackage{algorithm}
\usepackage{algorithmic}
\usepackage{graphicx}
\usepackage{textcomp}
\usepackage{xcolor}
\usepackage[numbers]{natbib}

\usepackage{hyperref}
\usepackage{mathtools}
\usepackage{amsthm}
\usepackage{subfigure}
\usepackage{booktabs}

\theoremstyle{plain}
\newtheorem{theorem}{Theorem}[section]

\theoremstyle{definition}
\newtheorem{definition}[theorem]{Definition}

\theoremstyle{remark}

\def\BibTeX{{\rm B\kern-.05em{\sc i\kern-.025em b}\kern-.08em
    T\kern-.1667em\lower.7ex\hbox{E}\kern-.125emX}}
\begin{document}

\title{High-dimensional Multi-objective Bayesian Optimization with Learned Variable Interactions\\
}

\author{\IEEEauthorblockN{1\textsuperscript{st} Hongyan Wang*}
\IEEEauthorblockA{\textit{School of Vehicle and Mobility} \\
\textit{Tsinghua University}\\
Beijing, China \\
hongyan-wang@mail.tsinghua.edu.cn}
\and
\IEEEauthorblockN{2\textsuperscript{nd} Jiayu Huang}
\IEEEauthorblockA{\textit{Faculty of Engineering} \\
\textit{The Hong Kong Polytechnic University}\\
Hong Kong, China \\
jiayu1999.huang@connect.polyu.hk}
\and
\IEEEauthorblockN{3\textsuperscript{rd} Haotian Zheng}
\IEEEauthorblockA{\textit{School of Vehicle and Mobility} \\
\textit{Tsinghua University}\\
Beijing, China \\
zhenght21@mails.tsinghua.edu.cn}
\and
\IEEEauthorblockN{4\textsuperscript{th} Xin Gao}
\IEEEauthorblockA{\textit{School of Vehicle and Mobility} \\
\textit{Tsinghua University}\\
Beijing, China \\
gaoxin97@mail.tsinghua.edu.cn}
\and
\IEEEauthorblockN{5\textsuperscript{th} Chi Ding}
\IEEEauthorblockA{\textit{School of Vehicle and Mobility} \\
\textit{Tsinghua University}\\
Beijing, China \\
dc22@mails.tsinghua.edu.cn}
\and
\IEEEauthorblockN{6\textsuperscript{th} Ying Liu}
\IEEEauthorblockA{\textit{School of Vehicle and Mobility} \\
\textit{Tsinghua University}\\
Beijing, China \\
seuliuy@hotmail.com}

\and
\IEEEauthorblockN{7\textsuperscript{th} Xia Wang}
\IEEEauthorblockA{\textit{School of Vehicle and Mobility} \\
\textit{Tsinghua University}\\
Beijing, China \\
wang-xia@mail.tsinghua.edu.cn}

\and
\IEEEauthorblockN{8\textsuperscript{th} Qing Xu}
\IEEEauthorblockA{\textit{School of Vehicle and Mobility} \\
\textit{Tsinghua University}\\
Beijing, China \\
qingxu@tsinghua.edu.cn}
\and
\IEEEauthorblockN{9\textsuperscript{th} Keqiang Li*}
\IEEEauthorblockA{\textit{School of Vehicle and Mobility} \\
\textit{Tsinghua University}\\
Beijing, China \\
likq@tsinghua.edu.cn}
}

\maketitle

\begin{abstract}
Multi-objective Bayesian optimization (MOBO) is effective in identifying the Pareto fronts for expensive black-box problems. However, most current MOBO approaches are limited to low-dimensional decision space due to its exponential sampling complexity. This paper presents decision variable interaction analysis-based MOBO, ViaMOBO, a generic framework for expensive multi-objective problems with high-dimensional decision space. The key idea of ViaMOBO is that it utilizes a variable interaction analysis model to determine whether the decision space can be completely or partially divided, and then performs local Bayesian optimization in the divided decision subspaces. Through the variable analysis model, it can be derived whether the objectives in black-box problems are separable, partially separable, or non-separable based on the potential independent or interdependent relationships among decision variables without any strong assumptions. We compare ViaMOBO with the state-of-the-art MOBO methods on both synthetic and real-world benchmarks. The experimental results demonstrate that ViaMOBO outperforms other related MOBO baselines in approximating the Pareto front of high-dimensional expensive multi-objective problems.

\end{abstract}

\begin{IEEEkeywords}
Multi-objective Bayesian optimization (MOBO), Variable Interaction Analysis, High-dimensional Space, Expensive multi-objective Problems.
\end{IEEEkeywords}

\section{Introduction}
Multi-objective problems (MOPs) are often encountered in various experimental design problems, such as optimization in robotics \cite{Schneider2013}, analog circuit design \cite{Wenlong2018}, aerospace engineering \cite{He2025,he2026}, and hyper-parameter tuning in neural networks \cite{Snoek12} etc. These problems are challenging to approximate the Pareto fronts (i.e., trade-offs among different objectives) since they are usually black-box problems with expensive evaluation costs, especially in high-dimensional space. 

Multi-objective Bayesian optimization (MOBO) ensures global convergence within limited computational cost, utilizing a probabilistic surrogate model (usually a Gaussian process surrogate) to approximate the original MOPs, and exploits an acquisition function to locate the next most promising points to be candidates for evaluation.

Many researches have focused on improving the performance of MOBO on trading off the relationship among multiple objectives in black-box problems, mainly including \textit{EGO-based methods, hypervolume improvement (HVI)-based methods} and \textit{predictive entropy-based methods}, etc. \textit{EGO-based methods} such as ParEGO \cite{Knowles2005}, SMS-EGO \cite{Ponweiser2008} and MOEA/D-EGO \cite{Zhang2010}, are extensions of EGO \cite{Jones1998}. These EGO-based approaches ensures the global convergence for MOPs. \textit{HVI-based methods} such as PEHVI \cite{YangPESB19}, hypervolume scalarizariion approach \cite{ZhangG2020}, qEHVI \cite{DaultonBB2020}, qNEHVI \cite{qnehvi2021} mainly aim to improve the hypervolume calculation efficiency in solving MOPs. \textit{Entropy-based methods} such as PESMO \cite{Hernandez-Lobato2016}, PESMOC \cite{pesmoc2019}, MESMO \cite{BelakariaDD2019}, MESMOC \cite{mesmoc2020}, PFES \cite{SuzukiTTSK2020} and MF-OSEMO \cite{BelakariaDD20} maximize the information gain of the Pareto optimum to select candidates for evaluation. Further MOBO studies mainly focus on uncertainty reduction in surrogate model PAL \cite{PAL2013}, improving the performance of scalarization \cite{PariaKP19,ZhangG2020}, solution diversity \cite{konakovic2020diversity} and acquisition function \cite{usemo20}. Although MOBO approaches have made great advances in approximating the Pareto fronts of black-box multi-objective problems, most of them are limited to low-dimensional decision variable space due to the exponential sampling complexity with decision variable dimensions \cite{Kandasamy2015HDB,wangICML2017a}. The performances of these methods are compromised greatly over high-dimensional decision space. 

For high-dimensional cases, there are few research concerned MOBO, ReMO \cite{QianY2017} and MORBO \cite{MORBO2021}. ReMO exploits random embedding \cite{Kandasamy2015HDB,wangICML2017a} to decompose the high-dimensional decision space into several subspaces with low dimensions to slove the black-box problems in low-dimensional space. MORBO uses local model with trust region \cite{ErikssonPGTP19} to perform local Bayesian optimization to search diverse solutions. However, the above methods manifest the following limitations. Firstly, ReMO is too hypothetical on the decision space. On one hand, it assumes that only few dimensions are effective while most dimensions do not affect the objectives. On the other hand, it assumes the same effective dimensions for all objectives in MOPs, which is obviously irrational for most optimization problems. Secondly, although without any strong assumption, MORBO aims at reducing the cubic time complexity with the number of observed data points, rather than emphasizing the high decision dimensionality and ignoring the sample efficiency of acquisition function. Nonparametric regression is challenging because of the curse of dimensionality \cite{daglib0035701}, i.e., computational complexity depending exponentially in dimension.  
Besides, commonly used optimization heuristics for acquisition function also requires exponential complexity with dimension, leading to high cost and low optimization efficiency. 

In real-world high-dimensional expensive MOPs, the interaction relationship may exist among decision variables and could jointly influence the performance space. The fluctuation of one variable may cause the change of the other variables, and thus change the objective values. Take hyper-parameter tunning in neural networks as an example, the learning rate and dropout rate may influence the model performance simultaneously. However, current MOBO ignore considering this relationship, which results in severely decreased performance. According to whether there's interaction relationship among decision variables, high-dimensional MOPs can be categoried into \textit{separable} and \textit{non-separable problems}. For separable problems, it is desirable to partition the decision space into disjoint subspaces by learning variable interaction relationship to reduce the sampling complexity and improve optimization efficiency, especially for high-dimensional problems. However, how to learn the interaction is a great challenge. 

For expensive high-dimensional separable MOPs, we present a MOBO framework, ViaMOBO, based on decision Variable Interaction Learning to alleviate the curse of dimensionality. ViaMOBO first determines whether a MOP is separable by learning the interaction among decision variables, and then divides the decision space into disjoint subspaces, with each subspace containing interdependent variables. To avoid real expensive function evaluation, ViaMOBO exploits a binary classifier to predict the magnitude relationship between two objective values of two decision variable vectors. In this way, ViaMOBO perform optimization for MOPs in low-dimensional decision sub-spaces, ensuring sample efficiency of acquisition function and avoiding the curse of dimensionality. Then, ViaMOBO selects a batch of candidates incorporating virtual derivate observations to avoid over-exploration of the search space as much as possible. Finally, trust region is used to avoid exponential complexity of observed points. Our contributions are as follows:
\begin{itemize}
\item A general MOBO framework, ViaMOBO, is presented, in which an additive kernel structure is first introduced into multi-objective settings, and the sample efficiency of the acquisition function in MOBO is considered. 

\item A decision variable interaction learning strategy is developed with the following properties: (a) the separability of high-dimensional MOPs is determined by learning the interaction relationships among decision variables; (b) expensive function evaluations are avoided through the use of a binary classifier for interaction learning.

\item The effectiveness of ViaMOBO is verified by comparisons with state-of-the-art MOBO methods on both synthetic and the real-world expensive problems. The experimental results demonstrate that ViaMOBO significantly outperforms the baselines.
\end{itemize}

\section{Background}
\subsection{Preliminaries}

\subsubsection{Bayesian Optimization (BO)}
\label{bo}

BO method uses a Gaussian process (GP) surrogate to model the objective, meanwhile using an acquisition function to recommend candidate solutions. Given a single-objective problem: $\mathbf{x}^{\ast} = arg\min_{\mathbf{x}\in X}f(\mathbf{x}), X\subseteq \mathbb{R}^D$, 
$f(\mathbf{x})$ is approximated with GP, i.e., $f(\mathbf{x})\sim \mathcal{GP}(m(\mathbf{x}),k(\mathbf{x,x'}))$, where $m(\mathbf{x})$, $k(\mathbf{x,x'})$ are the mean and kernel function, respectively. Given the noisy observations $\mathcal{D}_{1:t}=\{(x_i,y_i)\}\mid y_i=f(x_i)+\epsilon_i,i=1,\dots,t\}$, where $\epsilon\sim \mathcal{N}(0,\mathbf{\sigma}_{noise}^2)$, the predictive model is: $\mathcal{P}(\mathbf{y}_{t+1}\mid \mathcal{D}_{1:t},\mathbf{x}_{t+1})=\mathcal{N}(\mu_{t}(\mathbf{x}_{t+1}),\sigma_{t}^2(\mathbf{x}_{t+1}))$. $\mu_t(\mathbf{x}_{t+1})=\mathbf{k}^T(K+\mathbf{\sigma}_{noise}^2I)^{-1}\mathbf{y}_{1:t}$ and $\sigma_{t}^2(\mathbf{x}_{t+1}) = \mathbf{k}(\mathbf{x}_{t+1},\mathbf{x}_{t+1})-\mathbf{k}^T(K+\mathbf{\sigma}_{noise}^2I)^{-1}\mathbf{k}$ are the mean and variance function, respectively, and $\mathbf{k} = [\mathbf{k}(\mathbf{x}_{t+1},\mathbf{x}_{1}),\mathbf{k}(\mathbf{x}_{t+1},\mathbf{x}_{2}),\dots,\mathbf{k}(\mathbf{x}_{t+1},\mathbf{x}_{t})]$. Then, the acquisition function balancing exploitation (minimization of $\mu_t(\mathbf{x})$) and exploration (maximization of $\sigma_t(\mathbf{x})$) is built to search for the most promising candidates. 
 
\subsubsection{Separable and Non-Separable Problems}
\label{S-NS-Fs}
\begin{definition}
\label{variable-interact}
Two decision variables $x_i$ and $x_j$ are interacting (interdependent variables), if there exist $x, a_1, a_2, b_1$ and $b_2$ meeting \cite{ChenWYT10,xiaoliangma2016}
\begin{equation}
\begin{aligned}
& f(x)|_{x_i=a_2, x_j=b_1} < f(x)|_{x_i=a_1, x_j=b_1} \wedge \\
& f(x)|_{x_i=a_2, x_j=b_2} < f(x)|_{x_i=a_1, x_j=b_2},
\end{aligned}
\end{equation}
where 
\begin{equation}
\begin{aligned}
f(x)|_{x_i=a_2, x_j=b_1} \triangleq f(x_1,\dots,x_{i-1}, a_2,\dots, x_{j-1}, b_1,\dots, x_D).
\end{aligned}
\end{equation}
\end{definition}

\begin{definition}
\label{single-additive}
A function $f(x):\mathbb{R}^n\to \mathbb{R}$ is \textit{partially separable} if it is a sum of sub-functions $f(x)=\sum_{i=1}^{M}f_i(x_i)$, and each one depends on a group of interdependent decision variables. The function $f$ is called \textit{completely additively separable} or \textit{fully separable} function if all sub-functions are 1-$D$. $f(x)$ is non-separable if it is neither partially separable nor fully separable. \cite{BouzarkounaAD11}
\end{definition}

\begin{definition}
\label{MOP-sep}
A MOP is a separable multi-objective problem if every objective is fully separable. Otherwise, $F(x)$ is a non-separable MOP. \cite{wfg2006}
\end{definition}

For convenience, hereafter, we refer to both partially and fully separable function as separable functions in this paper, since under both cases, the decision variables can be decomposed to groups. The global optimum of a separable problem can be obtained by considering each dimension (fully separable) or each group of dimensions (partially separable) of variable in turn, independently of one another \cite{wfg2006}. For a MOP, this means that we can consider one dimension (fully separable) or a group of dimensions (partially separable) at one time to find at least some points on the Pareto front, which makes it easier for optimizer to approximate the Pareto front than non-separable MOPs. 

\subsection{High-dimensional Bayesian Optimization}
 
Various assumptions on the decision and objective/performance space have been made to extend BO to high ($\geq10$) dimensions \cite{Vanilla2024}. Random embedding-based methods, REMBO \cite{Wang2013HBO}, SI-BO \cite{Djolonga2013}, ALEBO \cite{LethamCRB2020}, PCA-BO \cite{RaponiWBBD20} assume that there are important and unimportant dimensions, and only consider the important dimensions of variables. Instead of directly optimizing in the original high-dimensional decision space, MS-UCB \cite{Huang2019HBO} performs acquisition function maximization over discretized low-dimensional subspaces embedded into the full space. By Assuming an additive structure for the function, ADD-GP-UCB \cite{Kandasamy2015HDB} reduces the computation exponential complexity in dimension when using global optimization heuristics to maximize the acquisition function. PP-GP-UCB \cite{LiKPS16}generalizes the additive assumption and the low-dimensional assumption to a projected-additive assumption to handle a broader class of functions. Batch-ADD-GP-UCB \cite{wangICML2017a} extends ADD-GP-UCB to sample a batch of candidate solutions. G-Add-GP-UCB \cite{RollandSBC18} extends GP-UCB by removing the restriction that the variable subsets must be mutually disjoint, thereby allowing additive structures with overlapping subsets.

TuRBO \cite{ErikssonPGTP19} uses trust regions to construct a group of local models and searches global optimum across these local models. However, most of these methods make too strong hypotheses about the decision or objective spaces, and they are confined to high-dimensional single-objective problems. MORBO \cite{MORBO2021} improves TuRBO not only by selecting trust region center in a coordinate fashion to improve convergence but also chooses new candidates by collaboratively optimized a shared hypervolume-based global utility. However, the trust region-based TuRBO and MORBO targets to address the challenge in fitting a global surrogate model when the function is heterogeneous, while failing to overcome the problem that the search space increases exponentially with dimension, which make global solver for acquisition function. To address this, single-objective random embedding with high dimensions has been scaled to high-dimensional MOPs that assumes only few dimensions influence the objective values \cite{QianY2017}. However, similar to other random embedding methods, ReMO assumes too ideal structure of the real-world black-box problems. Although other related approaches \cite{why2022,why2024} effectively address high-dimensional problems, the partitioning of the decision space involves elements of randomness and subjectivity.

\section{The proposed ViaMOBO}
\begin{algorithm*}[h]
\small
\caption{Framework of ViaMOBO}
\label{ViaMOBO}
\begin{algorithmic}[1]
    \STATE {\bfseries Input:} An MOP $F(x):\mathbb{R}^D\to \mathbb{R}^M$, maximum function evaluations $maxEvals$, sample budget $T$
    \STATE {\bfseries Output:} Approximate Pareto front $PF^*$
    \STATE Draw initial Sobol points $X_{init}$ and evaluate initial points $X_{init}$ with original MOP $F(x)$\ \ \ \ \ \ $\triangleright$ \textit{Initialization}
    \STATE Generate training and testing data given $D_0$ 
    \FOR{$m=1,\dots,M$}
    \STATE train SVM model $h_i(\cdot)$ for the $i$th objective $f_i(x)$
    \STATE Set $x$ defined in Eqn.(\ref{vars}) as the decision vector that maximizes the $i$th objective $f_i(x)$
    \STATE Perturb $x$ at the $i$th and $j$th dimension with random value to obtain $x_i', x_j'$ and $x_{ij}'$, where $i,j\in [1,D]\wedge i\neq j$
    \STATE Predict the magnitude relationship of the objective values of $x_i, x_i', x_j'$ and $x_{ij}'$ with the trained SVM model $h_i(\cdot)$
    \IF{$[f(x_i'),f(x_j')]$ dominates or is dominated by $[f(x),f(x_{ij}'')]$} 
    \STATE the $i$th decision variable interacts with the $j$th decision variable
    \ELSE
    \STATE the $i$th decision variable does not interact with the $j$th decision variable
    \ENDIF
    \STATE Learn all the interacting decision variables $\Omega_i$ of $f_i(x)$ based on transitivity defined in \textit{Def}.\ref{multiple-variable-interact}
    \ENDFOR\ \ \ \  $\triangleright$ \textit{Variable interaction learning}
    \STATE Learn the interdependent decision variables $D = \cup_j^N \Omega_j$ of $F(x)$ based on \textit{Def}.\ref{multiple-variable-interact-MOP} to judge whether $F(x)$ is separable
    \IF{$F(x)$ is non-separable}
    \STATE optimize $F(x)$ with other high-dimensional MOBO approach (e.g., MORBO \cite{MORBO2021})
    \ELSE
    \WHILE{$\textit{evals} \leq \textit{MaxEvals}$} \label{beginalg}
    \STATE Fit a local model for each objective $f_i(x)$, and obtain the additive structure of $F(x)$\ \ \ \ \ \ $\triangleright$ \textit{Additive kernel structure}
    \STATE Select $q$ candidates using the defined acquisition function (e.g., UCB, EI)\ \ \ \ \ \ $\triangleright$ \textit{Batch candidate sampling}
    \STATE  Evaluate candidates on $F(x)$ and obtain new observations
    \ENDWHILE \label{endalg}
    \ENDIF
\end{algorithmic}
\end{algorithm*}
This section introduces ViaMOBO, a multi-objective BO method with decision variable interaction based on theoretical analysis. As shown in Algorithm \ref{ViaMOBO}, to improve the sampling efficiency of acquisition function, ViaMOBO learns whether a MOP is separable by analyzing the variable interaction or not. Unlike random embedding-based methods \cite{Kandasamy2015HDB,wangICML2017a}, ViaMOBO makes no assumption on the structure of black-box problems, but deduces whether it is separable given prior observations or not. If the MOP is non-separable, we recommend to use other high-dimensional MOBO method, and we strongly recommend MORBO \cite{MORBO2021} here due to its high efficiency. For a separable MOP, ViaMOBO inferences the additive GPs and fits a local model of each objective in MOP in each lower dimensional decision sub-space. Then, multiple candidates are recommended by optimizing the acquisition function, and then evaluated by the true objective.

\subsection{Re-definition of Separable MOP}
\label{prob-redefine}
\textit{Def}.\ref{single-additive} gives the partially, fully separable and non-separable functions in single-objective cases. Hereafter, we refer to both partially and fully separable function as separable functions, since under both cases, the decision variables can be decomposed to groups. Given this, we redefine \textit{Def}.\ref{MOP-sep} for multi-objective cases as following. 
\begin{definition}
\label{MOP-sep-redefine}
A MOP is a separable multi-objective problem if every objective is separable (partially or fully separable). Otherwise, $F(x)$ is a non-separable MOP if all the objective are non-separable. 
\end{definition} 

\begin{definition}
\label{multiple-variable-interact}
If two decision variables $x_i$ and $x_j$ are interacting, and $x_k$ interacts with one of the two variables, then all these three variable are interacting. 
\end{definition}

Above definition indicates that there's transitivity among this characteristic and it can be extended to a case of infinite number of decision variable dimensions. This definition is necessary for MOPs due to the possible conflicting objectives. 

\begin{definition}
\label{multiple-variable-interact-MOP}
Given a separable MOP $F(x)=(f_1(x),\dots,f_M(x)):\mathbb{R}^D\to \mathbb{R}^M$, and the sub-functions of $f_i(x)=\sum_{j=1}^{M_i}f_j(x), x\in \Omega_i$ where $\Omega_i$ is the interacting variable partition of $f_i(x)$, then the interacting variable partition of $F(x)$ is $\cup_{i=1}^{M} \Omega_i$.
\end{definition}

\subsection{Variable Interaction Learning}
Based on \textit{Def}.\ref{variable-interact}, \textit{Def}.\ref{MOP-sep-redefine} and \textit{Def}.\ref{multiple-variable-interact}, we can find that if the interacting variables are obtained, then all the decision dimensions can be decomposed and then whether the objective is separable can be determined or not. Therefore, the key point is the variable interaction learning. For convenience, we re-write the decision vectors as below.
\begin{equation}
\label{vars}
\centering
\begin{aligned}
& x=(x_1,\dots, x_{i-1}, x_i=a_1,\dots, x_{j-1}, x_j=b_1, \dots, x_D)  \\
& x_i'=(x_1,\dots,x_{i-1}, x_i=a_2,\dots, x_{j-1}, x_j=b_1,\dots, x_D) \\
& x_j'=(x_1,\dots,x_{i-1}, x_i=a_1,\dots, x_{j-1}, x_j=b_2,\dots, x_D) \\
& x_{ij}''=(x_1,\dots,x_{i-1}, x_i=a_2,\dots, x_{j-1}, x_j=b_2,\dots, x_D) \\
\end{aligned}
\end{equation}
Re-examining \textit{Def}.\ref{variable-interact}, it indicates that if there exists a decision vector $x$ whose $i$th and $j$th variable ($a_1$ and $b_1$) can be substituted with values $a_2$ and $b_2$ to generate a strong dominance relationship between $[f(x_i'),f(x_j')]$ and $[f(x),f(x_{ij}'')]$ is established, then $x_i$ and $x_j$ are interacting. However, the function evaluations of many real-world MOPs are so expensive that it is unacceptable to cost function evaluations on additional objective value calculation $f(x_i'),f(x_j'), f(x_{ij}'')$.

To avoid the expensive function calculation of $x_i',x_j'\text{and}\ x_{ij}''$, we use a binary classifier (e.g., SVM) to learn the magnitude relationship of function values instead of evaluating $x_i',x_j'\text{and}\ x_{ij}''$ by using the true objective function. With initialized data points $\{X_{init}, F(x_{init})|F(x_{init})=(f_1(x),f_2(x),\dots, f_M(x))\}$, we generate training and test dataset of each objective $f_i(x)$ in $F(x)$ by concatenating two decision variables vectors $x_1, x_2$ into one vector, denoted as $x_1\oplus x_2$. If $f_i(x_1)<f_i(x_2)$, then the label is set to positive while negative if otherwise. To make the training and test dataset balanced, we regard $x_1\oplus x_2$ and $x_2 \oplus x_1$ as two different data. With the labeled samples, a binary classifier (e.g., supported Vector Machine (SVM)) is trained to learn the relationship of the objective values of two decision variable vectors. The training and test data ratio is $70\%$ and $30\%$. In this paper, we adopt SVM because it can maintain satisfactory classification performance in sparsely sampled regions while offering both parametric and non-parametric flexibility.

Specifically, to determine whether two dimensions in decision space of a MOP $F(x):\mathbb{R}^D\to \mathbb{R}^M$ are interacting or not, we first use \textit{Def}.\ref{variable-interact} on the observed decision variable $x$ that maximizes the $i$th objective $f_i(x)$ to separately learn the interacting variables of all $M$ objectives in MOPs. When learning, we perturb $x$ at the $i$th and $j$th dimension with random value in problem bounds to obtain $x_i', x_j', \text{and}\ x_{ij}'$. Then the trained SVM predicts the relationship $f(x_i'), f_(x_j')$ and $f_(x_{ij}')$ to compare whether there's a dominance relationship between $[f(x_i'),f(x_j')]$ and $[f(x),f(x_{ij}'')]$. Then, \textit{Def}.\ref{multiple-variable-interact} is used to learning all the possible interacting variables. Finally, we use \textit{Def}.\ref{multiple-variable-interact-MOP} to deduce the interdependent variable partition $D = \cup_j^N \Omega_j$ of $F(x)$.

\subsection{Additive Kernel for MOPs}
If the MOP $F(x):\mathbb{R}^D\to \mathbb{R}^M$ is separable, then 
\begin{equation}
\begin{aligned}
F(x)=(\sum_{i=1}^{N}f_1^i(x), \sum_{i=1}^{N}f_2^i(x),\dots, \sum_{i=1}^{N}f_M^i(x)),
\end{aligned}
\end{equation}
where $D = \cup_j^N \Omega_j$ and $\Omega_i \cap \Omega_j=\varnothing, i,j\in[1,\dots,N], i\neq j$. Similar to ADD-GP-UCB \cite{Kandasamy2015HDB}, we call $\Omega_i$ disjoint sub-groups. For each sub-objective $f_i(x)$ in $F(x)$, $f_i(x)$ is single-objective and additive,
\begin{equation}
\begin{aligned}
f_i(x)=f^{(1)}(x^{(1)})+f^{(2)}(x^{(2)})+\dots+f^{(N)}(x^{(N)}),
\end{aligned}
\end{equation}
where $x^{(j)}\in \Omega_i$. Following single-objective ADD-GP-UCB \cite{Kandasamy2015HDB}, we assume $f^{(j)}\sim GP(\mu_i^{(j)(x)}, k_j^{(j)}(x^{(i)},x^{(j)'}))$, then $f_i(x)\sim GP(\mu_i(x),k_i(x,x'))$ in noiseless case. The mean $\mu_i(x)$ and kernel function $k_i(x,x')$ are 
\begin{equation}
\label{add-mu-sigma}
\begin{aligned}
\mu_i(x) &= \mu^{(1)}(x^{(1)})+ \dots +\mu^{(N)}(x^{(N)})\\
k_i(x,x')&=k^{(1)}(x^{(1)},x^{(1)'}) + \dots + k^{(M)}(x^{(M)},x^{(M)'}).
\end{aligned}
\end{equation}
The sub-kernel $k^{(j)}(x^{(j)},x^{(j)'})$ is defined in the decision sub-space $\Omega_i$. Then, we can inference the posterior distribution of the component $f_{t+1}^{(j)}(x)$ given the observed data $\{(X_t, F_t(x))|X=\{x_1,\dots, x_D \}, F_t(x) = (f_1^t(x), f_2^t(x),\dots,f_M^t(x))\}$,  is defined as the following. 
\begin{equation}
\begin{aligned}
&f_{t+1}^{(j)}(x) \sim GP(\mu^{(j)}_{t+1}(x^{(j)}),k^{(j)}_{t+1}(x^{(j)},x^{(j)'})),\\
&\mu^{(j)}_{t+1}(x^{(j)})=k^{(j)}(x_{t+1}^{(j)},X^{(j)}k(X,X)^{-1}) \mathbf{f_i},\\
&k^{(j)}_{t+1}(x^{(j)},x^{(j)'})=k^{(j)}(x_{t+1}^{(j)}, x_{t+1}^{(j)'})\\
&-k^{(j)}(x_{t+1}^{(j)}, X^{(j)})k(X,X)^{-1}k^{(j)}(X,x^{(j)})
\end{aligned}
\end{equation} 
and thus $f_{t+1}(x) \sim GP(\mu_{t+1}(x),k_{t+1}(x,x'))$. More details about single-objective additive kernel can be found in ADD-GP-UCB \cite{Kandasamy2015HDB}. 

In this way, given the observed data $\{(X_t, F_t(x))|X=\{x_1,\dots, x_D \}, F_t(x) = (f_1^t(x), f_2^t(x),\dots,f_M^t(x))\}$, we can obtain the posterior of $F(x)$ as the following. 
\begin{equation}
\begin{aligned}
&F(x)\sim (GP(\mu^{t+1}_1(x),k_1^{t+1}(x,x')),\\
&GP(\mu^{t+1}_2(x),k_2^{t+1}(x,x')),\dots,GP(\mu^{t+1}_M(x),k_M^{t+1}(x,x')).
\end{aligned}
\end{equation}

\subsection{Batch candidate sampling}
As discussed in \ref{S-NS-Fs}, for a separable MOP, we can consider one dimension (fully separable) or a group of dimensions (partially separable) a time to find at least some points on the Pareto front. Therefore, after we learn the additive structure and inference the additive structure of posterior distribution of MOP, we can acquire candidate points by solving the acquisition functions in lower decision sub-spaces $\Omega_i$, such as EI \cite{Jones1998}, UCB \cite{Kandasamy2015HDB} and EHVI \cite{qnehvi2021}.

\textbf{Sampling with UCB.} ADD-GP-UCB shows an additive UCB acquisition function based on the additive kernel, i.e., $\varphi_t(x) = \mu_{t-1}(x)+\beta_t^{1/2}\sum_{j=1}^{N}\sigma_{t-1}^{(j)}(x^{(j)})$, where $\mu(x)$ and $\sigma^2(x)$ ($k(x,x')$) are defined in Eqn (\ref{add-mu-sigma}). The additive acquisition function can be maximized by maximizing each component in decision subspace $\Omega_i$, which is much easier than optimizing it in $D$-dimensional decision space. Based on single-objective ADD-GP-UCB, we can deduce the multi-objective UCB as the following.
\begin{equation}
\begin{aligned}
A_t(x) = (\varphi_1^t(x), \varphi_2^t(x),\dots,\varphi_1^M(x)),
\end{aligned}
\end{equation}
where each $\varphi_i^t(x),i\in[1,\dots,M]$ is an additive UCB. To obtain a batch of candidates, we use $q$UCB strategy \cite{DaultonBB2020} to scalarize all the $M$ additive UCBs.
\begin{figure*}[t]
\centering
\subfigure{
\label{10d_dtlz2}
\includegraphics[width=0.63\columnwidth]{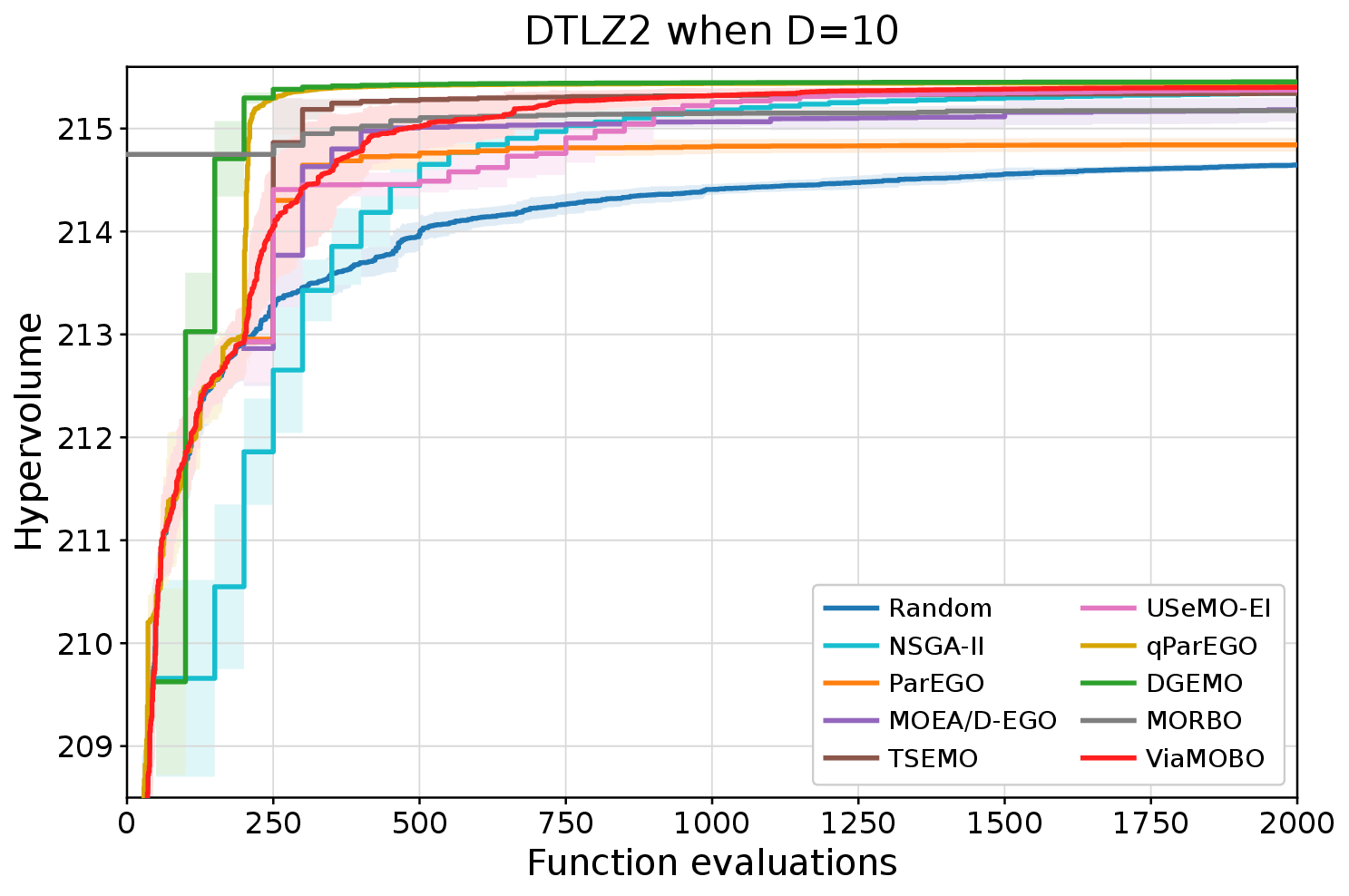}
}
\centering
\subfigure{
\label{30d_dtlz2}
\includegraphics[width=0.63\columnwidth]{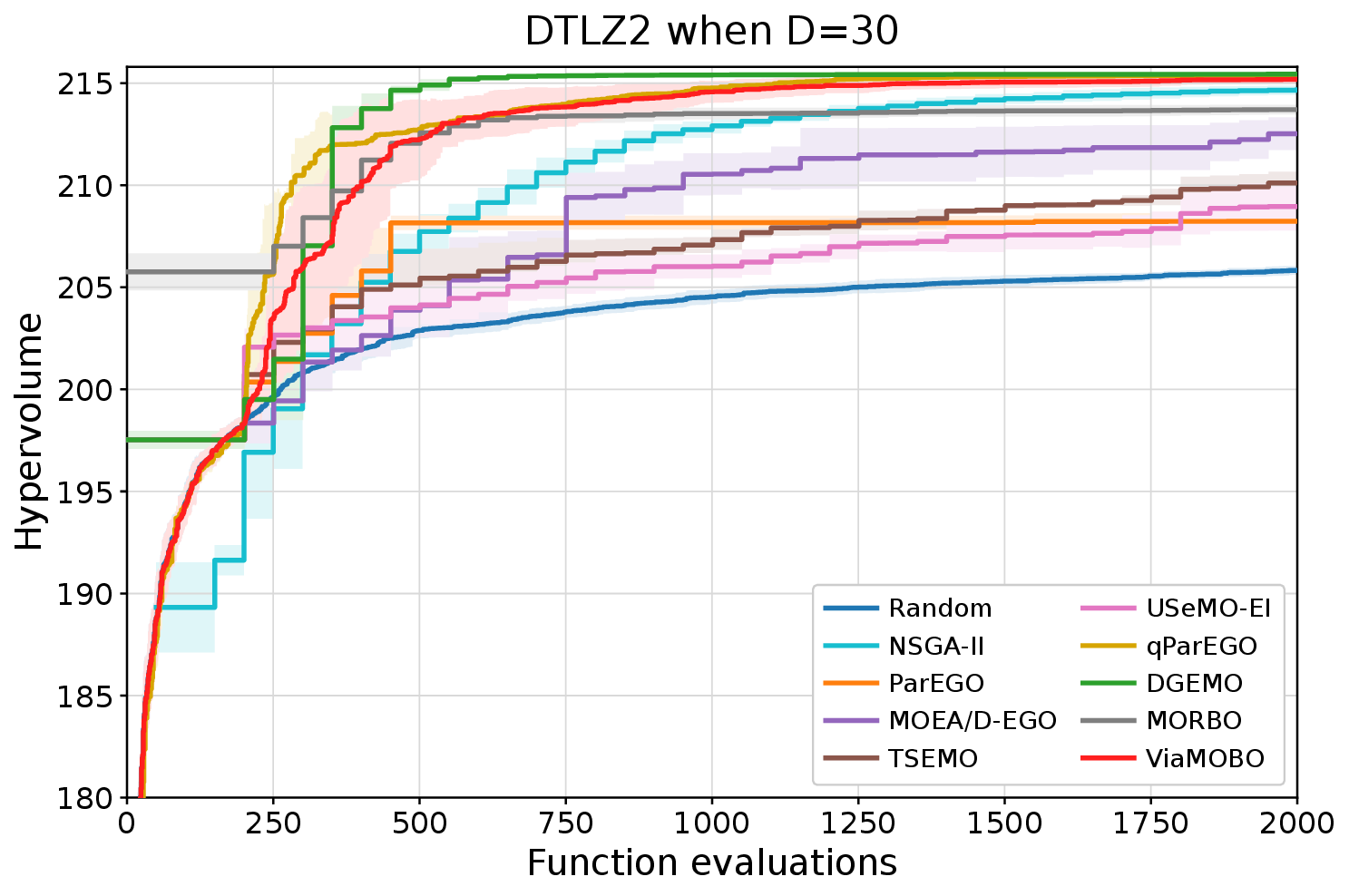}
}
\centering
\subfigure
{
\label{100d_dtlz2}
\includegraphics[width=0.63\columnwidth]{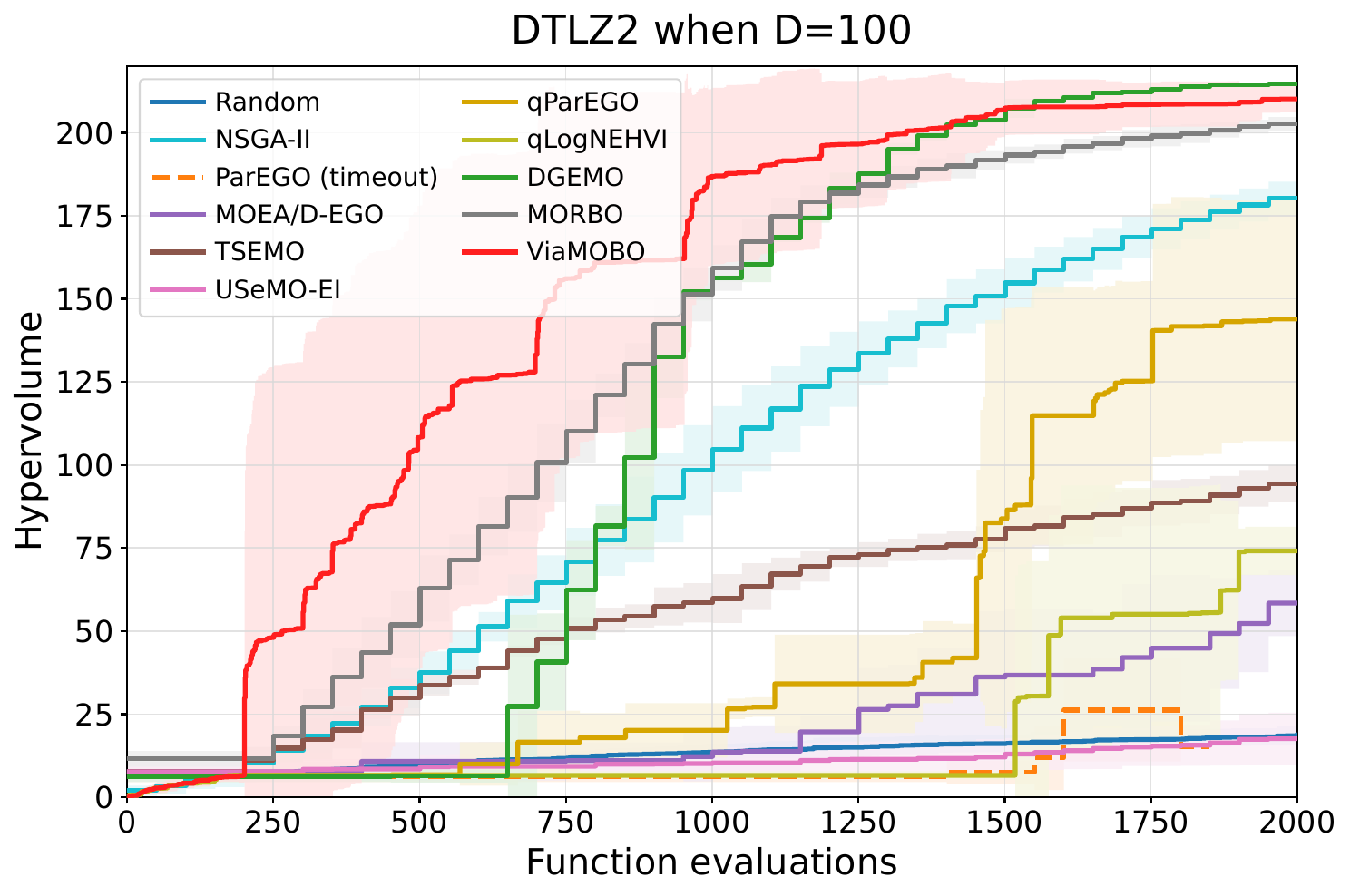}
}
\caption{(Left) Illustration of all the related methods on synthetic problem with relative high $10D$ decision dimensions. (Medium)  Illustration of all the related methods on synthetic problem with relative high $30D$ decision dimensions. (Right) Illustration of all the related methods on synthetic problem with higher $100D$ decision dimensions.}
\label{dtlz2}
\end{figure*}

\textbf{Sampling with EI and EHVI.}Although EI and EHVI can not be directly deduced as additive EI in multi-objective cases even based on our additive structure, we can acquire it in $|\Omega_i|$-dimensional decision space, which obviously improves the optimization efficiency for global optimization heuristics \cite{Kandasamy2015HDB,wangICML2017a}. We use $q$EI and EHVI \cite{qnehvi2021} ($q$EHVI is out of memory in our experiments) as our sampling strategy in this paper.

To draw large batch sizes $q$ of candidates, we borrow the idea from MORBO\cite{MORBO2021} that uses Thompson sampling to obtain $q$ posteriors from GP, and optimizes the acquisition function group by group. However, it is undesirable to sequentially maximize the first group and then the next group. Instead, we select the $k$th candidate by maximizing the $k$th group $\Omega_k$ of decision space and all other groups we use values from the Pareto optimums from the previous observed points. After $N$ iterations we cycle through the groups if $q>N$. $N$ is the number of sub-groups of the decision space. To avoid over-exploration (boundary issue) \cite{OhGW18} for high-dimensional BO methods when optimizing acquisition, we use the virtual derivative sign observations \cite{SiivolaVVGA18}. 

\section{Experiments and Results}

\subsection{Experimental Settings} 
To validate the effectiveness of ViaMOBO, it is compared with NSGA-II\footnote{https://github.com/anyoptimization/pymoo}\cite{NSGAII2002}, ParEGO\footnotemark[\value{footnote}]\cite{Knowles2005}, 
MOEA/D-EGO\footnotemark[\value{footnote}]\cite{Zhang2010}, TSEMO\footnote{https://github.com/yunshengtian/DGEMO}\cite{bradford2018efficient}, USeMO-EI\footnotemark[\value{footnote}]\cite{usemo20}, DGEMO\footnotemark[\value{footnote}]\cite{konakovic2020diversity}, $q$ParEGO \cite{DaultonBB2020}, $q$qLogNEHVI\cite{qlogehvi2023}, MORBO \cite{MORBO2021}, 
and random search with Sobol---a quasi-random baseline \cite{qnehvi2021}.$q$-series methods, MORBO and ViaMOBO experiments are conducted using BoTorch \footnote{https://botorch.org/} \cite{botorch20}. Among the EHVI-family methods, we select $q$LogNEHVI as a representative baseline due to its numerically stable logarithmic reformulation of the acquisition function. All methods are run for 10 replications, and the initial data points used to build GP are quasi-random for each replication. We use the same hypar-prameters for all the related methods. Hypervolume (HV) is used to evaluate the related performances. The acquisition functions used in this paper are EI \cite{Jones1998}, UCB \cite{Kandasamy2015HDB} and EHVI \cite{qnehvi2021}. All of them are Monte-Carlo acquisition functions \cite{botorch20}. 

The benchmarks used in this paper include synthetic functions and real-world benchmarks. Specifically, we first measure the performances of all methods on the synthetic function, separable DTLZ2 with $D=10$, $D=30$ and $D\geq100$ with $3$ objectives. Then, all the methods are evaluated on three real-world problems, including two aerodynamic shape-optimization\cite{aye2023airfoil} problems and trajectory planning\cite{MORBO2021}. Specifically, trajectory planning is with $D=60$ and exhibits strong sequential coupling because neighboring control points jointly determine the resulting trajectory. The airfoil design benchmark optimizes $20$ and $40$ shape parameters with three objectives—drag, lift, and geometric regularity—using XFOIL under fixed aerodynamic conditions. 
The reference point for HV calculation of is $\mathbf r_{min}=[6,6,6]$ and $\mathbf r_{\min}=(0.25,\;0.0,\;0.5)$ for $3$-objective DTLZ2, airfoil design problem, respectively.

All the related methods are initialized $200$ with Sobol points to construct the initial GP model. For all the synthetic problems, we use $2000$ maximum function evaluations, and 50 batch size for all the related methods. We find that these two methods even can not run when there is three objective. Therefore, we set the batch size to $5$ for these two methods in this paper. For ViaMOBO, we use $70\%$ and $30\%$ of the initial $200$ points to generate our training and test datasets. We set the perturbation of decision values to 100 in the normalized standard bounds $[0,1]$ of all the related problems to find if there is $a_2, b_2$ and to determine whether two variables are interacting. We use a constant-mean independent GP with an RBF kernel is utilized, and its hyperparameters are estimated by maximizing the marginal log-likelihood. This GP model is consistently applied to all BO baseline methods. For all the related methods, we use Monte-Carlo (MC) acquisition functions \cite{botorch20}, including EI, UCB and EHVI. Following MORBO \cite{MORBO2021}, we use the same number of quasi-MC samples and Fourier basis functions. For $q$ParEGO and $q$LogNEHVI, the acquisition function is optimized via L-BFGS-B using 20 random restarts. For MORBO and ViaMOBO, we use 4096 discrete points to optimize the related acquisition function for all problems. Details can be found in MORBO \cite{MORBO2021}. Besides, we use virtual deriviate information to overcome over-exploration in high-dimensional cases. 
\begin{table*}[t]
    \centering
    \caption{Hypervolume performance and computational cost on the DTLZ2 problems with 10, 30, and 100 decision variables. The hypervolume results are reported as mean $\pm$ standard deviation. HV@500 and HV@1000 denote the hypervolume values obtained after 500 and 1,000 function evaluations, respectively. AUC-HV denotes the normalized area under the hypervolume convergence curve. Time per batch is calculated over 36 optimization batches with a batch size of $q=50$. Relative runtime is normalized by the runtime of ViaMOBO at the corresponding dimensionality. All experiments are conducted on a server equipped with two Intel Xeon Platinum 8470Q CPUs (104 physical cores and 208 logical threads), 754 GiB of RAM, and an NVIDIA GeForce RTX 5090 GPU with 32 GB of memory.}
    \label{tab:dtlz2_combined_performance_runtime}
    \renewcommand{\arraystretch}{1.10}
    \setlength{\tabcolsep}{3.2pt}
    \resizebox{\textwidth}{!}{%
        \begin{tabular}{c l c c c c c c c}
            \hline
            Dim. & Algorithm & HV@500 & HV@1000 & Final HV & AUC-HV & Runtime per seed & Time per batch & Relative runtime \\
            \hline
            10 & Random & 213.975 & 214.409 & $214.646 \pm 0.033$ & 213.80 & 1.87 s & -- & $<0.001\times$ \\
            10 & NSGA-II & 214.652 & 215.181 & $215.347 \pm 0.014$ & 214.45 & 3.63 s & -- & $<0.001\times$ \\
            10 & ParEGO & 214.777 & 214.839 & $214.851 \pm 0.065$ & 214.59 & $17.92 \pm 2.03$ h & 1792 s & $11.1\times$ \\
            10 & MOEA/D-EGO & 215.015 & 215.065 & $215.184 \pm 0.113$ & 214.79 & $5.27 \pm 0.65$ h & 527 s & $3.3\times$ \\
            10 & TSEMO & 215.283 & 215.321 & $215.342 \pm 0.008$ & 215.04 & $1.68 \pm 0.47$ h & 168 s & $1.0\times$ \\
            10 & USeMO-EI & 214.489 & 215.259 & $215.373 \pm 0.020$ & 214.83 & $3.56 \pm 0.62$ h & 356 s & $2.2\times$ \\
            10 & qParEGO & 215.419 & 215.438 & $215.446 \pm 0.000$ & 215.40 & $5.49 \pm 0.55$ h & 549 s & $3.4\times$ \\
            10 &  qLogNEHVI & --  & --  & -- &-- & $>48$ h (timeout) & -- & -- \\
            10 & DGEMO & $\mathbf{215.428}$ & $\mathbf{215.445}$ & $\mathbf{215.454 \pm 0.001}$ & $\mathbf{215.16}$ & $9.13 \pm 0.55$ h & 913 s & $5.6\times$ \\
            10 & MORBO & 215.077 & 215.146 & $215.174 \pm 0.029$ & 215.08 & $14.46 \pm 3.25$ h & 1446 s & $8.9\times$ \\
            10 & ViaMOBO & 215.019 & 215.321 & $215.402 \pm 0.020$ & 214.61 & $1.62 \pm 0.37$ h & 162 s & $1.0\times$ \\
            \hline

            30 & Random & 202.885 & 204.536 & $205.837 \pm 0.207$ & 202.68 & 1.95 s & -- & $<0.001\times$ \\
            30 & NSGA-II & 207.729 & 212.909 & $214.674 \pm 0.217$ & 209.31 & 3.45 s & -- & $<0.001\times$ \\
            30 & ParEGO & 208.153 & 208.168 & $208.232 \pm 0.323$ & 206.47 & $17.25 \pm 0.99$ h & 1725 s & $10.0\times$ \\
            30 & MOEA/D-EGO & 203.874 & 210.534 & $212.524 \pm 0.654$ & 207.66 & $5.57 \pm 0.13$ h & 557 s & $3.2\times$ \\
            30 & TSEMO & 205.112 & 207.070 & $210.108 \pm 0.455$ & 206.22 & $3.69 \pm 0.43$ h & 369 s & $2.1\times$ \\
            30 & USeMO-EI & 203.997 & 206.015 & $208.959 \pm 0.957$ & 205.29 & $4.77 \pm 0.13$ h & 477 s & $2.8\times$ \\
            30 & qParEGO & 212.747 & 214.760 & $215.396 \pm 0.023$ & 213.95 & $6.42 \pm 0.26$ h & 642 s & $3.7\times$ \\
            30 & qLogNEHVI & --  & --  & -- &-- & $>48$ h (timeout) & -- & -- \\
            30 & DGEMO & $\mathbf{214.659}$ & $\mathbf{215.398}$ & $\mathbf{215.439 \pm 0.003}$ & $\mathbf{212.52}$ & $6.84 \pm 0.47$ h & 684 s & $4.0\times$ \\
            30 & MORBO & 212.052 & 213.511 & $213.704 \pm 0.235$ & 212.05 & $5.42 \pm 0.66$ h & 542 s & $3.1\times$ \\
            30 & ViaMOBO & 212.253 & 214.571 & $215.184 \pm 0.217$ & 210.97 & $1.73 \pm 0.62$ h & 173 s & $1.0\times$ \\
            \hline

            100 & Random & 9.785 & 13.524 & $18.790 \pm 2.741$ & 12.72 & 3.43 s & -- & $<0.001\times$ \\
            100 & NSGA-II & 32.829 & 98.413 & $180.236 \pm 4.822$ & 95.18 & 0.59 s & -- & $<0.001\times$ \\
            100 & ParEGO & 7.167 & 7.167 & $37.925 \pm 21.965$ & 13.97 & $>48$ h (timeout) & -- & $>12.0\times$ \\
            100 & MOEA/D-EGO & 10.786 & 12.248 & $58.452 \pm 8.142$ & 21.66 & $18.38 \pm 1.07$ h & 1838 s & $4.6\times$ \\
            100 & TSEMO & 30.010 & 58.495 & $94.433 \pm 4.487$ & 55.22 & $26.74 \pm 1.09$ h & 2674 s & $6.7\times$ \\
            100 & USeMO-EI & 7.230 & 8.334 & $15.660 \pm 7.131$ & 9.33 & $25.10 \pm 0.35$ h & 2510 s & $6.3\times$ \\
            100 & qParEGO & 6.806 & 20.123 & $143.954 \pm 36.732$ & 50.80 & $8.69 \pm 0.87$ h & 869 s & $2.2\times$ \\
            100 & qLogNEHVI & 6.584 & 6.584 & $74.132 \pm 5.916$ & 18.29 & $9.85 \pm 0.65$ h & 985 s & $2.5\times$ \\
            100 & DGEMO & 6.384 & 152.107 & $\mathbf{214.679 \pm 0.114}$ & 115.82 & $41.90 \pm 0.40$ h & 4190 s & $10.5\times$ \\
            100 & MORBO & 51.898 & 151.351 & $202.670 \pm 2.099$ & 127.110 & $2.836 \pm 1.341$ h & 283.6 s & $0.71\times$ \\
            100 & ViaMOBO & $\mathbf{108.291}$ & $\mathbf{186.882}$ & $210.205 \pm 3.789$ & $\mathbf{148.91}$ & $3.99 \pm 2.16$ h & 399 s & $1.0\times$ \\
            \hline
        \end{tabular}%
    }
\end{table*}

\subsection{Results on Synthetic Benchmark Problems} 
We first consider a separable synthetic problem, DTLZ2, a widely used test problem to study the performance of multi-objective methods. Strictly speaking, DTLZ2 is not separable, since optimizing one decision variable at a time will not identify all the global optimum. However, there are multiple global optimum and it is many-to-one, if optimizing along each coordinate direction is sufficient to attain a global optimum \cite{wfg2006}. Therefore, it is usually regarded as a separable MOP. Figure~\ref{dtlz2} presents the HV convergence curves on the 10-, 30-, and 100-dimensional DTLZ2 problems. Table~\ref{tab:dtlz2_combined_performance_runtime} shows HV$@$500, HV@1000, final HV and their computational costs on DTLZ2. The first two metrics characterize early- and intermediate-stage convergence, while the final HV represents the performance after 2,000 function evaluations\footnote{qLogNEHVI incurred substantial computational overhead due to Monte Carlo estimation, Pareto-set processing, and hypervolume partitioning for 10D and 3D 3-objective DTLZ2. Its runtime is not determined solely by the decision-space dimensionality; the larger non-dominated sets encountered in the low dimensional cases increased the partitioning cost. Under the given limit, only one 10D run completed successfully (HV$@$500, HV@1000, AUC-HV and final HV are 215.272, 215.272, 215.299, 215.16 using 35.46h, respectively). Therefore, qLogNEHVI is reported only for the 100D problem.}.


\begin{figure}[t]
\centering
\includegraphics[width=\columnwidth]{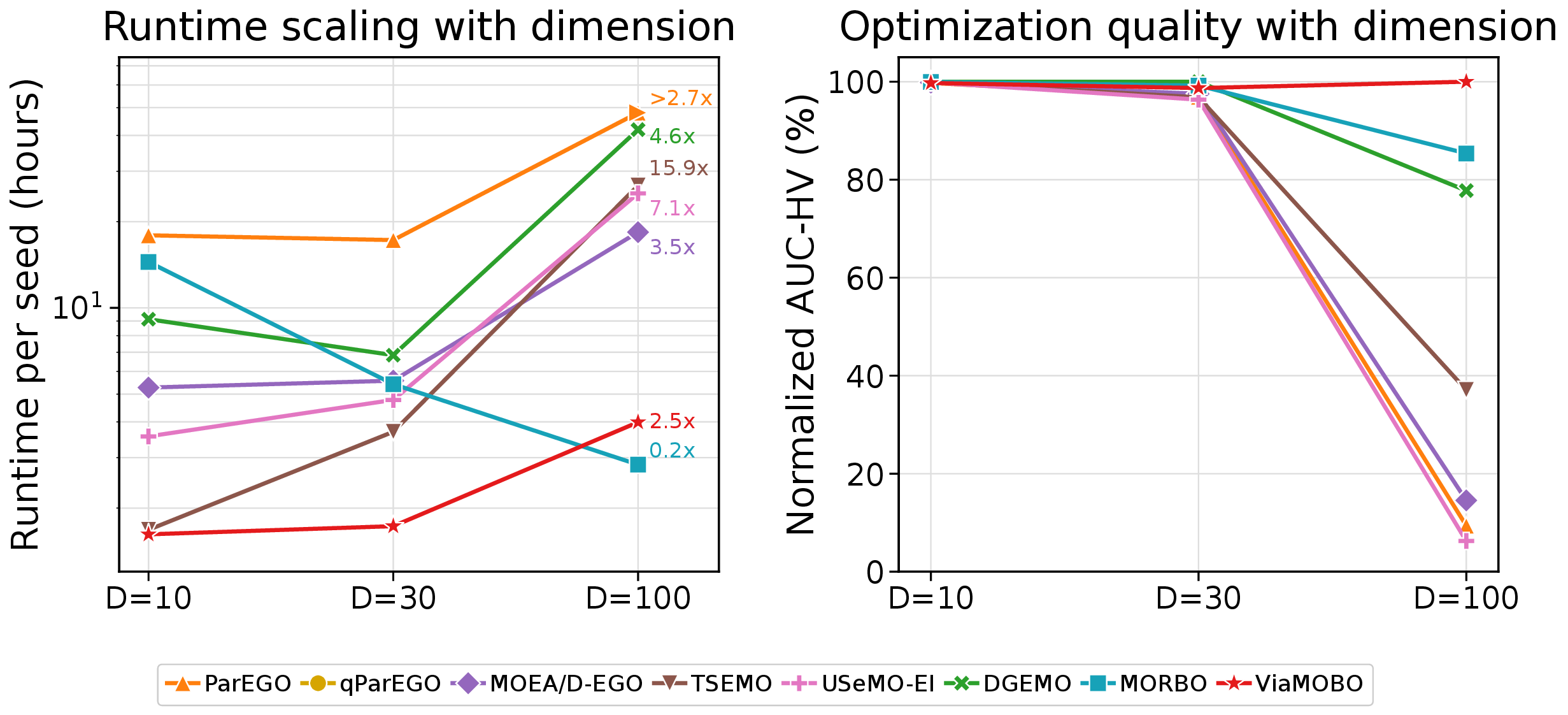}
\caption{(Left) The average runtime per seed on a logarithmic scale from 10D to 100D. (Right) AUC-HV normalized by the best result at each dimensionality. }
\label{time-hv-dtlz2}
\end{figure}

\begin{table*}[t]
    \centering
    \caption{Computational cost on the 20- and 40-dimensional Airfoil optimization problems. Runtime is reported as mean $\pm$ standard deviation. Time per batch is calculated using 36 optimization batches with $q=50$. Relative runtime is normalized by the runtime of ViaMOBO at the corresponding dimensionality.}
    \label{tab:airfoil_runtime}
    \renewcommand{\arraystretch}{1.10}
    \setlength{\tabcolsep}{3pt}
    \resizebox{0.8\textwidth}{!}{%
        \begin{tabular}{lccc|ccc}
            \toprule
            & \multicolumn{3}{c|}{20D}
            & \multicolumn{3}{c}{40D} \\
            \cmidrule(lr){2-4}\cmidrule(lr){5-7}
            Algorithm
            & Runtime/seed & Time/batch & Relative
            & Runtime/seed & Time/batch & Relative \\
            \midrule
            Random
            & $0.077 \pm 0.003$h & -- & $0.097\times$
            & $0.110 \pm 0.009$h & -- & $0.081\times$ \\

            NSGA-II
            & $0.082 \pm 0.003$h & -- & $0.103\times$
            & $0.104 \pm 0.004$h & -- & $0.077\times$ \\

            ParEGO
            & $18.193 \pm 1.691$h & 1819.3s & $22.76\times$
            & $>48$h (\textit{timeout}) & -- & $>35.47\times$ \\

            qParEGO
            & $6.704 \pm 0.413$h & 670.4s & $8.39\times$
            & $8.532 \pm 0.435$h & 853.2s & $6.31\times$ \\

            MOEA/D-EGO
            & $3.476 \pm 0.396$h & 347.6s & $4.35\times$
            & $6.394 \pm 1.039$h & 639.4s & $4.72\times$ \\

            TSEMO
            & $3.760 \pm 0.428$h & 376.0s & $4.70\times$
            & $8.156 \pm 1.286$h & 815.6s & $6.03\times$ \\

            USeMO-EI
            & $3.592 \pm 0.382$h & 359.2s & $4.49\times$
            & $7.867 \pm 0.618$h & 786.7s & $5.81\times$ \\

            DGEMO
            & $9.045 \pm 1.691$h & 904.5s & $11.32\times$
            & $18.328 \pm 4.590$h & 1832.8s & $13.54\times$ \\

            MORBO
            & $8.985 \pm 3.973$h & 898.5s & $11.24\times$
            & $9.371 \pm 5.133$h & 937.1s & $6.93\times$ \\

            ViaMOBO
            & $\mathbf{0.799 \pm 0.119}$h
            & $\mathbf{79.9}$s
            & $\mathbf{1.00\times}$
            & $\mathbf{1.353 \pm 0.174}$h
            & $\mathbf{135.3}$s
            & $\mathbf{1.00\times}$ \\
            \bottomrule
        \end{tabular}%
    }
\end{table*}
From Figure~\ref{dtlz2} and Table~\ref{tab:dtlz2_combined_performance_runtime}, we observe that on the 10-dimensional problem, DGEMO achieves the highest final HV, followed closely by qParEGO and ViaMOBO. The difference between DGEMO and ViaMOBO is only $0.024\%$, whereas DGEMO and qParEGO require approximately $5.6\times$ and $3.4\times$ the runtime of ViaMOBO, respectively. Therefore, ViaMOBO provides a near-optimal Pareto-front approximation at substantially lower computational cost. On the 30-dimensional problem, DGEMO obtains the highest HV@500, HV@1000, final HV, and AUC-HV. qParEGO also achieves a competitive final HV only approximately $0.020\%$ below DGEMO, but requires approximately $3.7\times$ the runtime of ViaMOBO. ViaMOBO is only $0.118\%$ below DGEMO, while requiring approximately one-quarter of its runtime. Other baselines require $2.1$--$10.0\times$ the runtime of ViaMOBO while producing lower final HV. On the 100-dimensional problem, ViaMOBO exhibits a clear early- and intermediate-stage convergence advantage, achieving the highest HV@500 and HV@1000 values. It also obtains the highest AUC-HV, which is approximately $28.6\%$ higher than DGEMO's. Although DGEMO ultimately achieves a final HV approximately $2.13\%$ higher than ViaMOBO, it requires approximately $10.5\times$ the runtime. MORBO's final HV and AUC-HV are approximately $3.58\%$ and $14.64\%$ lower than those of ViaMOBO, respectively. Therefore, although MORBO remains competitive in terms of final solution quality, ViaMOBO provides faster convergence and better overall performance under a limited evaluation budget. The observed MORBO runtime is $2.836\pm1.341$ hours per seed. qParEGO requires approximately $2.2\times$ the runtime of ViaMOBO but obtains a substantially lower final HV, indicating reduced effectiveness as the dimensionality increases. Figure~\ref{time-hv-dtlz2} further demonstrates that ViaMOBO exhibits a relatively moderate increase in computational cost while maintaining competitive optimization quality as the dimensionality increases.

Overall, ViaMOBO achieves near-optimal performance at low dimensionality and demonstrates an increasingly advantageous balance among convergence speed, final solution quality, and computational cost in high-dimensional decision spaces. 


\subsection{Results on Real-world Problems}
This section considers real-world problems, including 20- and 40-dimensional airfoil benchmark optimization problem and a bi-objective trajectory planning problem with $60$ decision variables \cite{MORBO2021}. We compare the related methods using 2000 total function evaluations with batch size $q=50$ at each iteration. 
\begin{figure}[t]
\centering
\subfigure
{
\label{fig:airfoil_hv}
\includegraphics[width=0.45\columnwidth]{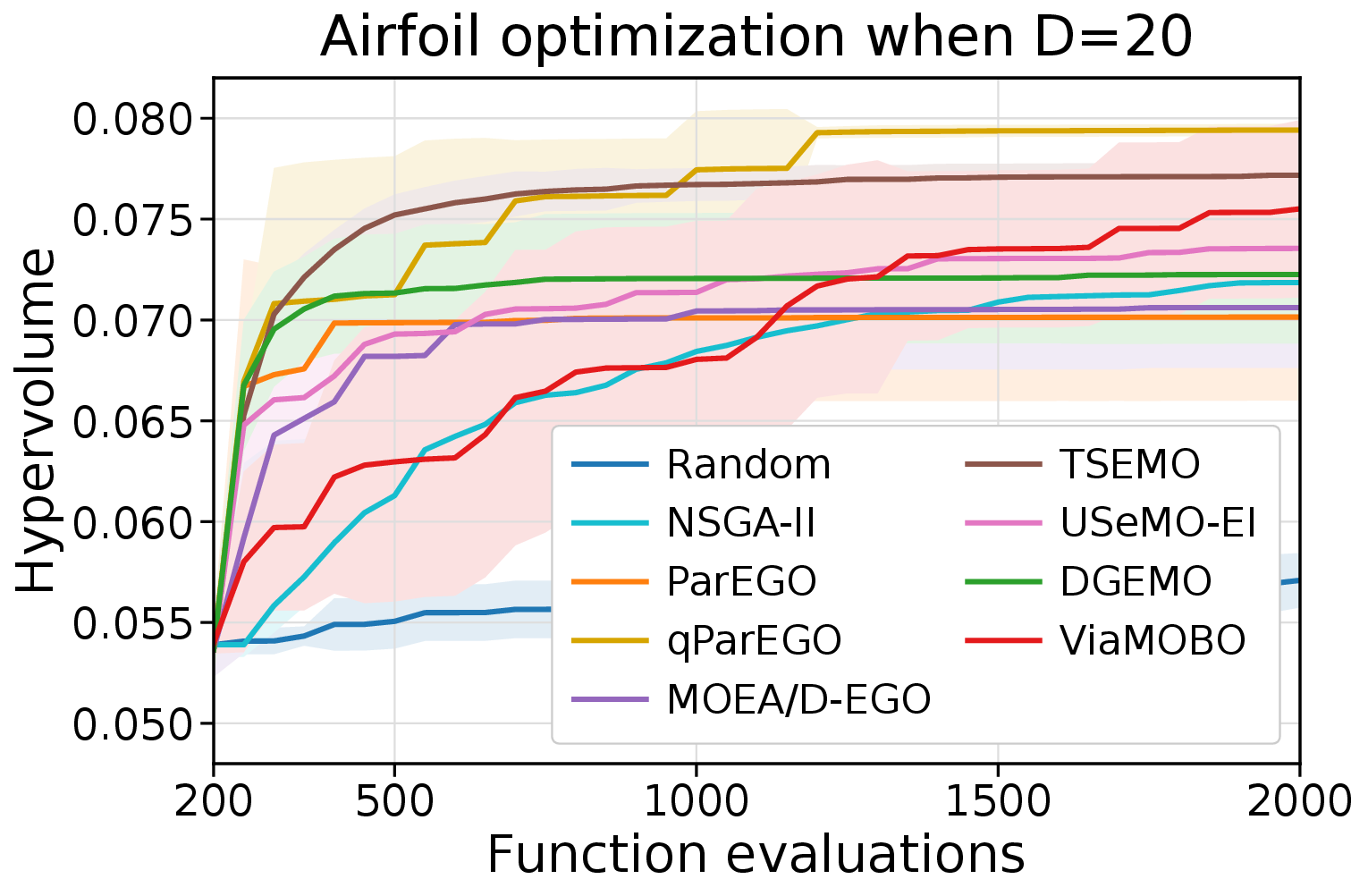}
}
\centering
\subfigure
{
\label{fig:airfoil_tradeoff}
\includegraphics[width=0.45\columnwidth]{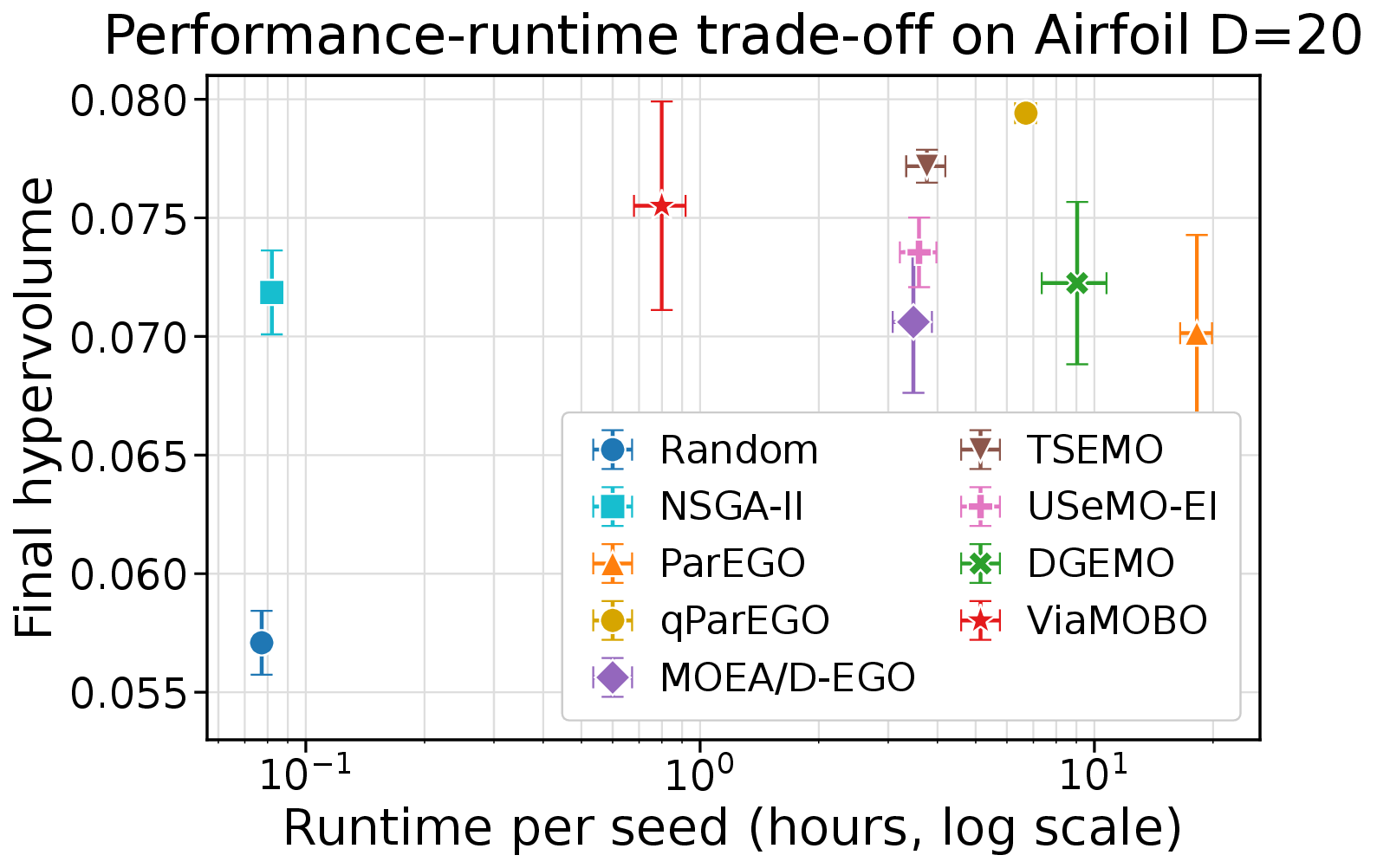}
}
\centering
\caption{\textbf{(Left)} The results of log HV difference and \textbf{(Right)} HV–runtime of all methods on airfoil problem with $20$ decision variables.}
\label{af}
\end{figure}

\subsubsection{20D Airfoil shape-optimization problem}
As shown in Figure~\(\ref{fig:airfoil_hv}\), TSEMO exhibits fast early-stage convergence on the 20-dimensional Airfoil problem, whereas qParEGO continues to improve and achieves the highest final HV. MORBO also demonstrates strong optimization performance. ViaMOBO converges relatively slowly during the early stage but improves consistently throughout the optimization process. Its final HV is approximately \(2.16\%\) lower than that of TSEMO and \(4.92\%\) lower than that of qParEGO, while remaining competitive with the other baselines. Table~\(\ref{tab:airfoil_runtime}\) and Figure~\(\ref{fig:airfoil_tradeoff}\) further illustrate the trade-off between optimization performance and computational cost. ViaMOBO requires only \(0.799\pm0.119\) hours per seed, making it the fastest surrogate-based method. Although TSEMO and qParEGO achieve higher final HV values, they require approximately \(4.70\times\) and \(8.39\times\) the runtime of ViaMOBO, respectively. MORBO requires \(8.985\pm3.973\) hours per completed seed, corresponding to \(11.24\times\) the runtime of ViaMOBO. DGEMO and ParEGO similarly require approximately \(11.32\times\) and \(22.76\times\) the runtime, respectively, without achieving better final performance than qParEGO or MORBO. Therefore, ViaMOBO provides a more favorable balance between solution quality and computational efficiency.
\begin{figure}[t]
\centering
\subfigure
{
\label{fig:airfoil40_hv}
\includegraphics[width=0.45\columnwidth]{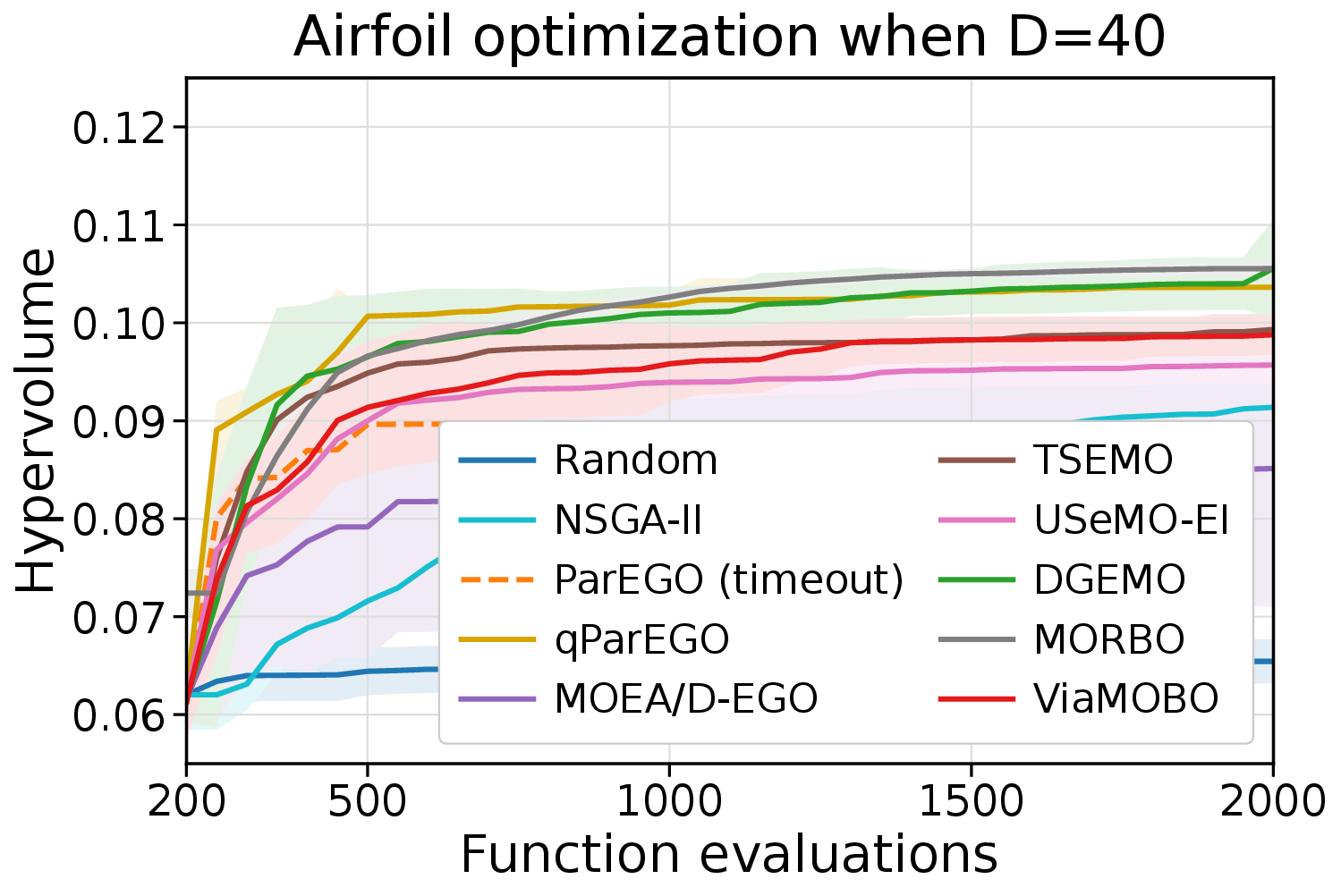}
}
\centering
\subfigure
{
\label{fig:airfoil40_tradeoff}
\includegraphics[width=0.45\columnwidth]{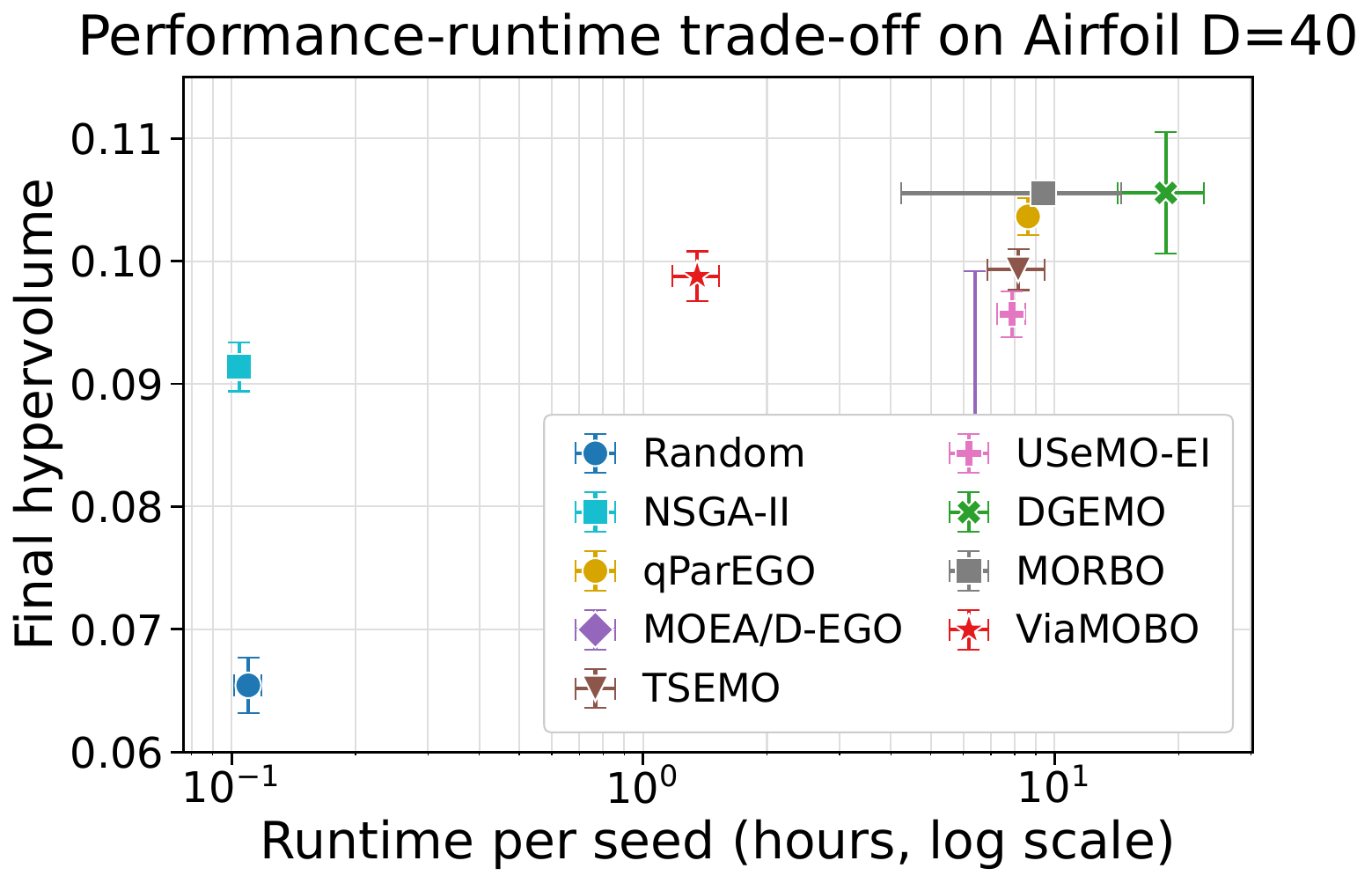}
}
\centering
\caption{\textbf{(Left)} The results of log HV difference and \textbf{(Right)} HV–runtime of all methods on airfoil problem with $20$ decision variables.}
\label{fig:airfoil40}
\end{figure}


Overall, qParEGO achieves the highest final solution quality, while MORBO and TSEMO also exhibit strong optimization performance. In contrast, ViaMOBO attains competitive final performance using only approximately \(21.3\%\), \(11.9\%\), and \(8.9\%\) of the runtimes required by TSEMO, qParEGO, and MORBO, respectively. These results demonstrate the suitability of ViaMOBO for high-dimensional practical optimization when computational resources are limited. 

\subsubsection{40D Airfoil shape-optimization problem}
As shown in Figure~\ref{fig:airfoil40} and Table~\ref{tab:airfoil_runtime}, MORBO achieves the highest final HV on the 40-dimensional Airfoil problem, but requires $9.371\pm5.133$ hours on average, approximately $6.92\times$ the runtime of ViaMOBO. In comparison, ViaMOBO attains a competitive final HV in only $1.353\pm0.174$ hours, making it the most computationally efficient surrogate-based method among the completed runs. Although MORBO, qParEGO, DGEMO, and TSEMO improve the final HV over ViaMOBO by approximately $6.84\%$, $4.92\%$, $6.31\%$, and $0.54\%$, respectively, they require about $6.92\times$, $6.37\times$, $13.54\times$, and $6.03\times$ more runtime. These results indicate that ViaMOBO provides a more favorable trade-off between Pareto-front quality and computational cost, particularly for high-dimensional expensive multi-objective optimization under limited computational budgets.

MORBO's computational cost is sensitive to its internal optimization trajectory. On the 100-dimensional problem, the trust regions remain relatively large, fewer invalid centers are encountered, and model fitting and candidate generation require only approximately $0.37$--$0.42$ and $0.59$--$1.53$ hours, respectively. By contrast, candidate generation alone requires approximately $8.58$--$12.01$ hours. Therefore, the shorter runtime observed in 100 dimensions reflects a less demanding search trajectory rather than improved scaling with dimensionality. This interpretation is further supported by its inferior optimization performance, with an AUC-HV of approximately $126.82$, compared with $148.91$ for ViaMOBO, together with a lower final HV. Accordingly, the reported MORBO runtime should be regarded as an empirical wall-clock measurement rather than evidence of reduced computational complexity at higher dimensions. A rigorous scalability comparison would require all dimensional settings to be evaluated under identical and isolated computational resources.
\begin{figure}[t]
\centering
\subfigure
{
\label{rover}
\includegraphics[width=0.45\columnwidth]{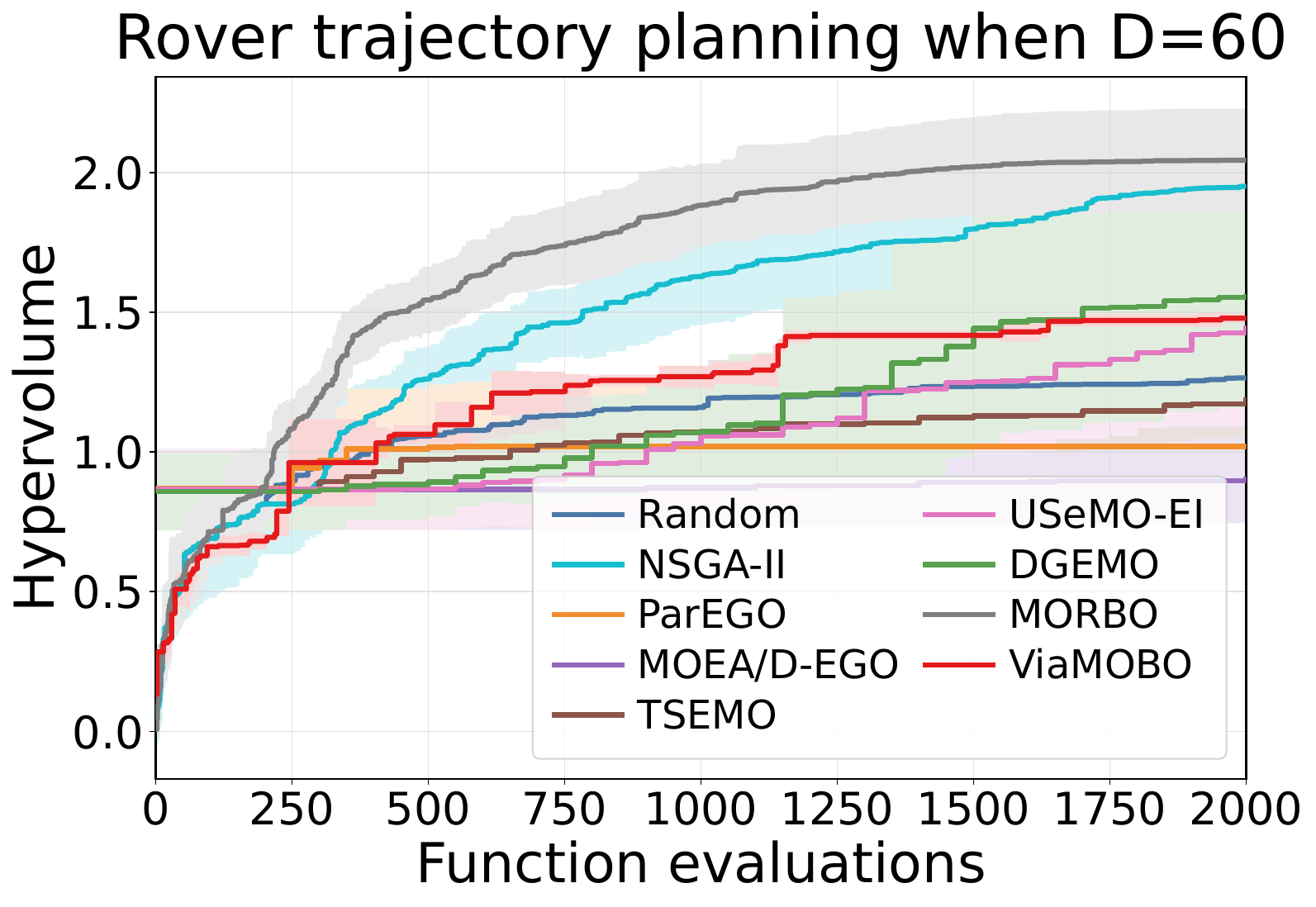}
}
\centering
\subfigure
{
\label{rover}
\includegraphics[width=0.45\columnwidth]{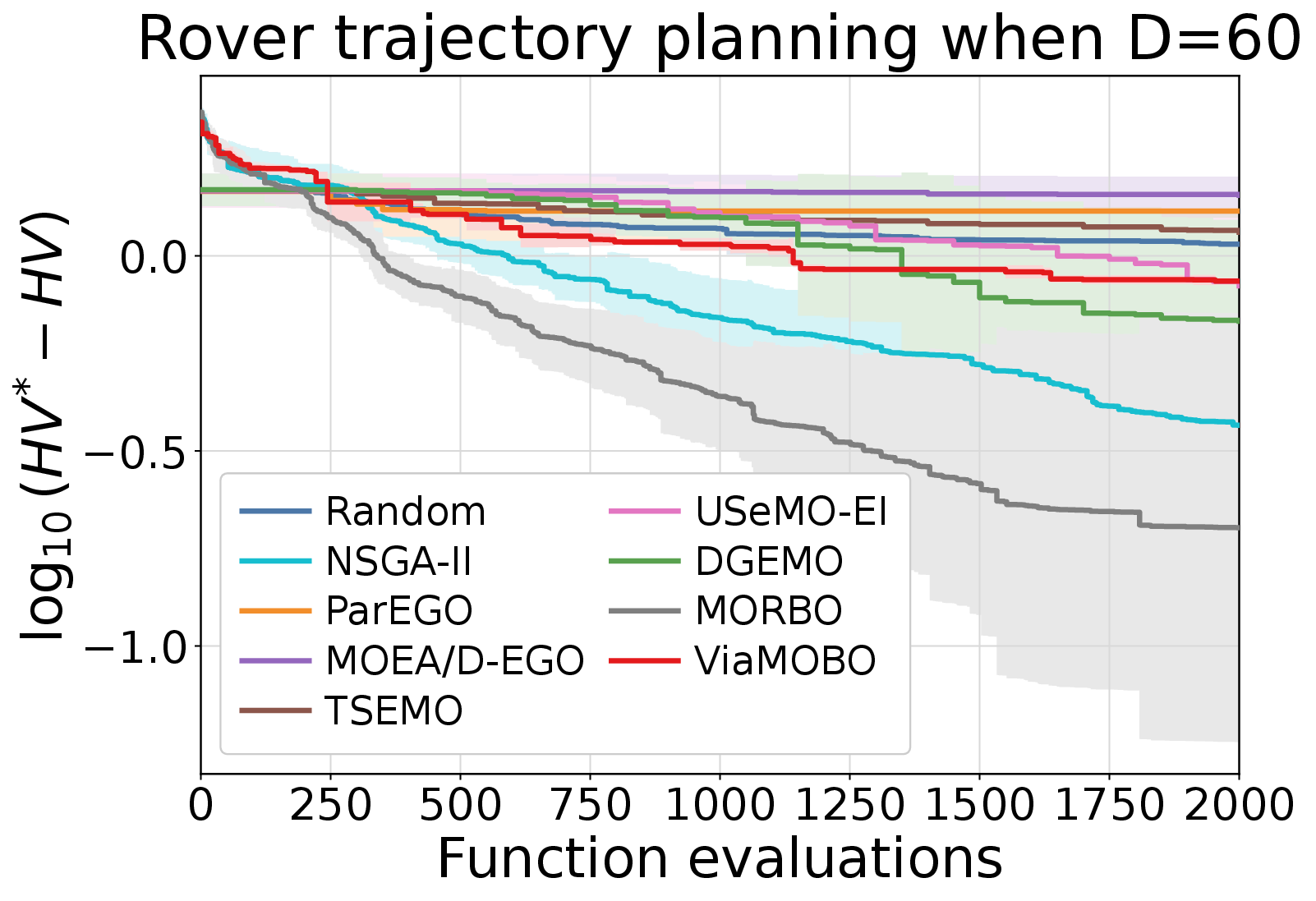}
}
\centering
\caption{\textbf{(Left)} The results of HVs and \textbf{(Right)} log HV differences of all methods on bi-objective trajectory planning with $60$ decision variables.}
\label{rover-all-hvs}
\end{figure}

\subsubsection{Trajectory planning problem} 
Figure \ref{rover-all-hvs} shows the HV and log-HV difference of the evaluated algorithms on the 60-dimensional Rover trajectory-planning problem. The $\log$-HV difference is $\log_{10}(HV^{*}-HV)$ where a lower value indicates better convergence toward the reference hypervolume. Figure \ref{rover-all-hvs}, we can see that MORBO and NSGA-II achieve the best performance, respectively. ViaMOBO outperforms Random, ParEGO, MOEA/D-EGO, TSEMO, and USeMO-EI and remaining comparable to DGEMO, but falling behind MORBO and NSGA-II. MORBO exhibits both the fastest decrease and the lowest final value, followed by NSGA-II. Although ViaMOBO continuously reduces the hypervolume gap, its convergence becomes slower during the middle and later stages. These results suggest that the variable-grouping mechanism of ViaMOBO may omit important cross-group dependencies on strongly coupled sequential problems. Consequently, the Rover result illustrates an applicability boundary of ViaMOBO: it is better suited to high-dimensional problems with identifiable group structure or relatively weak inter-group coupling than to trajectory-optimization tasks with strong continuous coupling.

\subsection{Ablation study}
\label{ablation}

Finally, to study the effectiveness of ViaMOBO, we first show the model accuracy of our binary classifier SVM, and then study the characteristics of ViaMOBO with different acquisition functions, i.e., EI, UCB and EHVI.
\begin{figure}[t]
\centering
\subfigure
{
\label{rover}
\includegraphics[width=0.45\columnwidth]{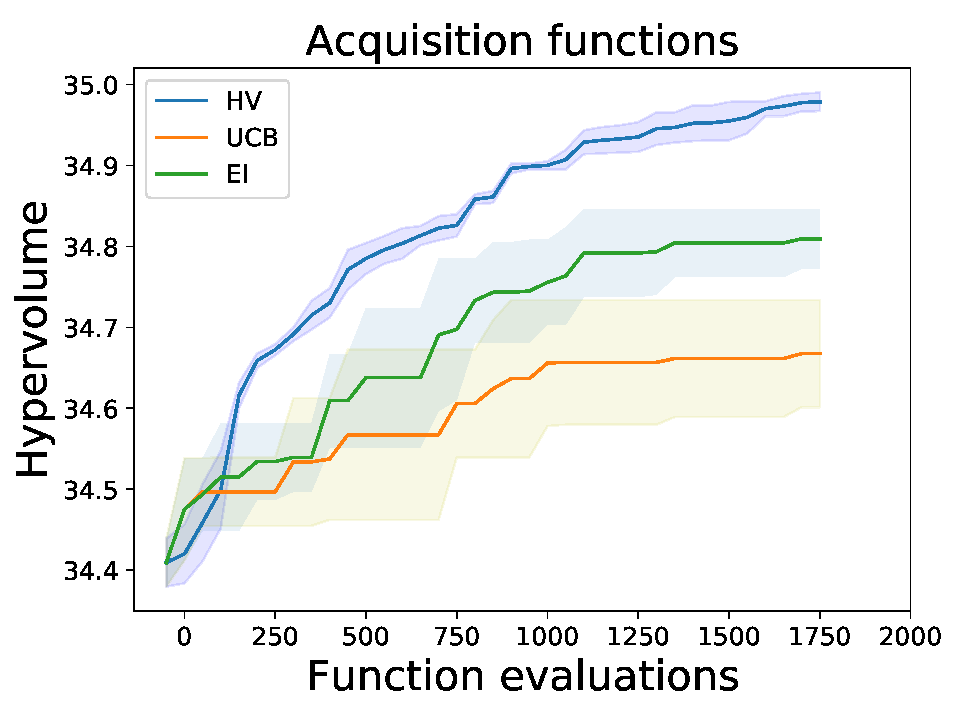}
}
\centering
\subfigure
{
\label{af-hvs}
\includegraphics[width=0.45\columnwidth]{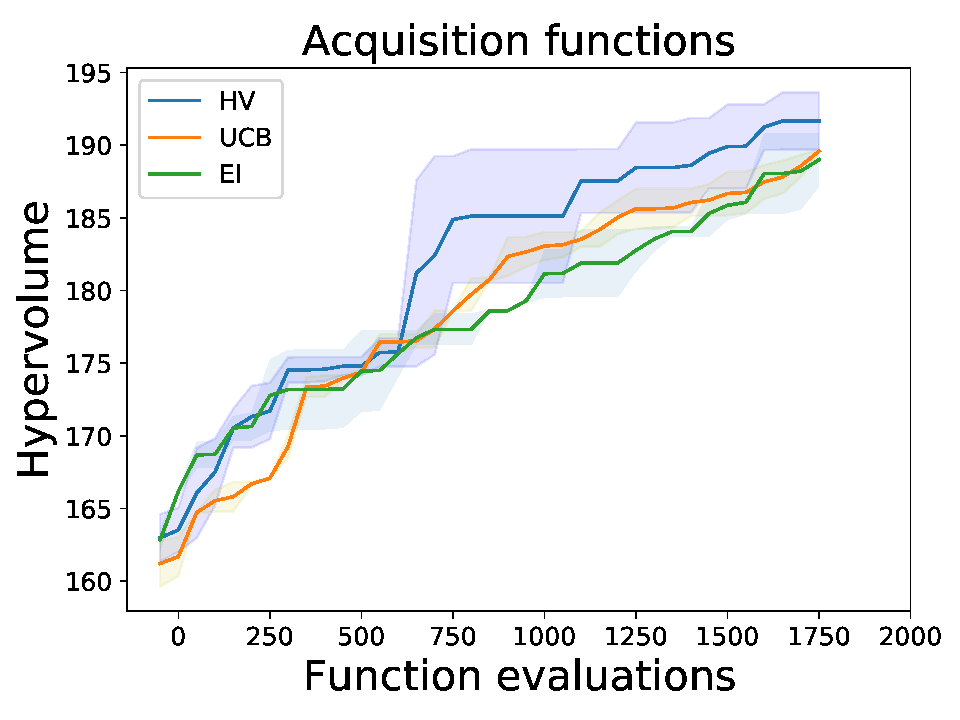}
}
\caption{The hypervolume performances of ViaMOBO with EI, UCB and EHVI on DTLZ2 with $10$ (Left) and $50$ (Right)decision variables. }
\label{af_comapre}
\end{figure}

\subsubsection{Accuracy of the binary classifier} We record the mean of accuracy of 10 runs for each objective of 3-objective DTLZ2. Table~\ref{tab:viamobo_grouping_accuracy} reports the variable-grouping accuracy of ViaMOBO on DTLZ2 with different dimensionalities. The overall grouping accuracy decreases from $90.68\%$ at $D=10$ to $73.77\%$ at $D=100$, indicating that identifying variable interactions becomes more challenging as the dimensionality increases. Nevertheless, ViaMOBO maintains an overall accuracy above $73\%$ even for the 100-dimensional problem, demonstrating that the grouping model remains effective in high-dimensional search spaces.
The final-iteration accuracy follows a similar trend, decreasing from $90.23\%$ at $D=10$ to $72.91\%$ at $D=100$. The difference between the overall and final-iteration accuracies is small for all dimensionalities, suggesting that the learned grouping structures remain relatively stable throughout the optimization process. The larger standard deviation at $D=100$ indicates increased variability across independent runs in higher-dimensional spaces. The $D=50$ results should be interpreted with caution because only two independent runs are currently available. 
\begin{table}[t]
    \centering
    \caption{Variable-grouping model accuracy of ViaMOBO on DTLZ2.}
    \label{tab:viamobo_grouping_accuracy}
    \renewcommand{\arraystretch}{1.12}
    \setlength{\tabcolsep}{5pt}
    \begin{tabular}{ccc}
        \toprule
        Dimension
        & Overall accuracy (\%)
        & Final accuracy (\%) \\
        \midrule
        10  & $90.68 \pm 0.26$ & $90.23 \pm 1.78$ \\
        30  & $81.71 \pm 0.63$ & $80.59 \pm 1.26$ \\
        50  & $78.61 \pm 1.04$ & $76.46 \pm 1.67$ \\
        100 & $73.77 \pm 1.32$ & $72.91 \pm 2.80$ \\
        \bottomrule
    \end{tabular}
\end{table}

\subsubsection{Performance of different acquisition functions} 
This section study the HV and computation cost performances of different acquisition functions in our proposed ViaMOBO framework, including EI, UCB and EHVI. 

\begin{figure}[t]
\centering
\subfigure
{
\label{3obj-af-hvs}
\includegraphics[width=0.45\columnwidth]{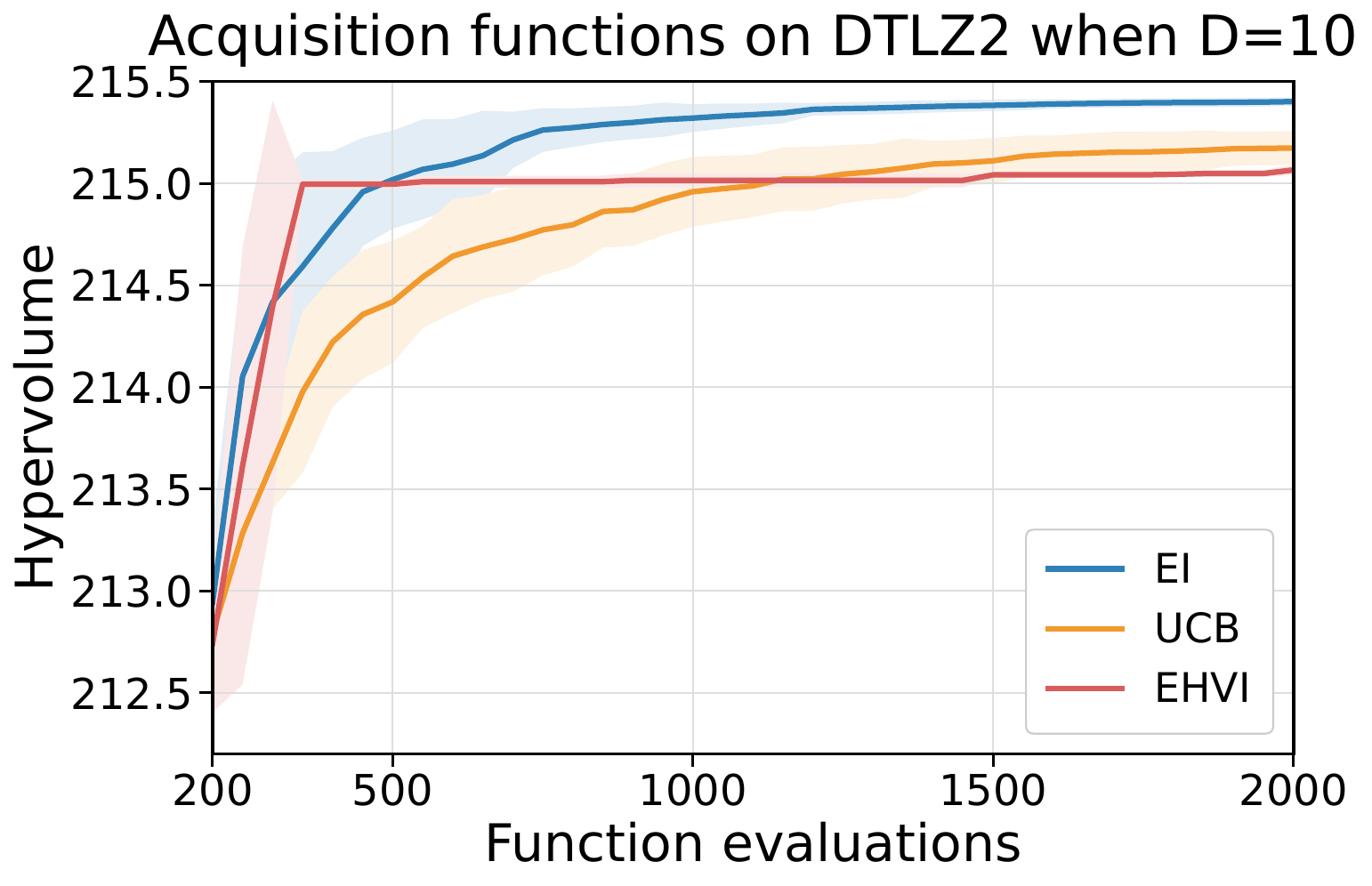}
}
\centering
\subfigure
{
\label{3obj-wall-time}
\includegraphics[width=0.45\columnwidth]{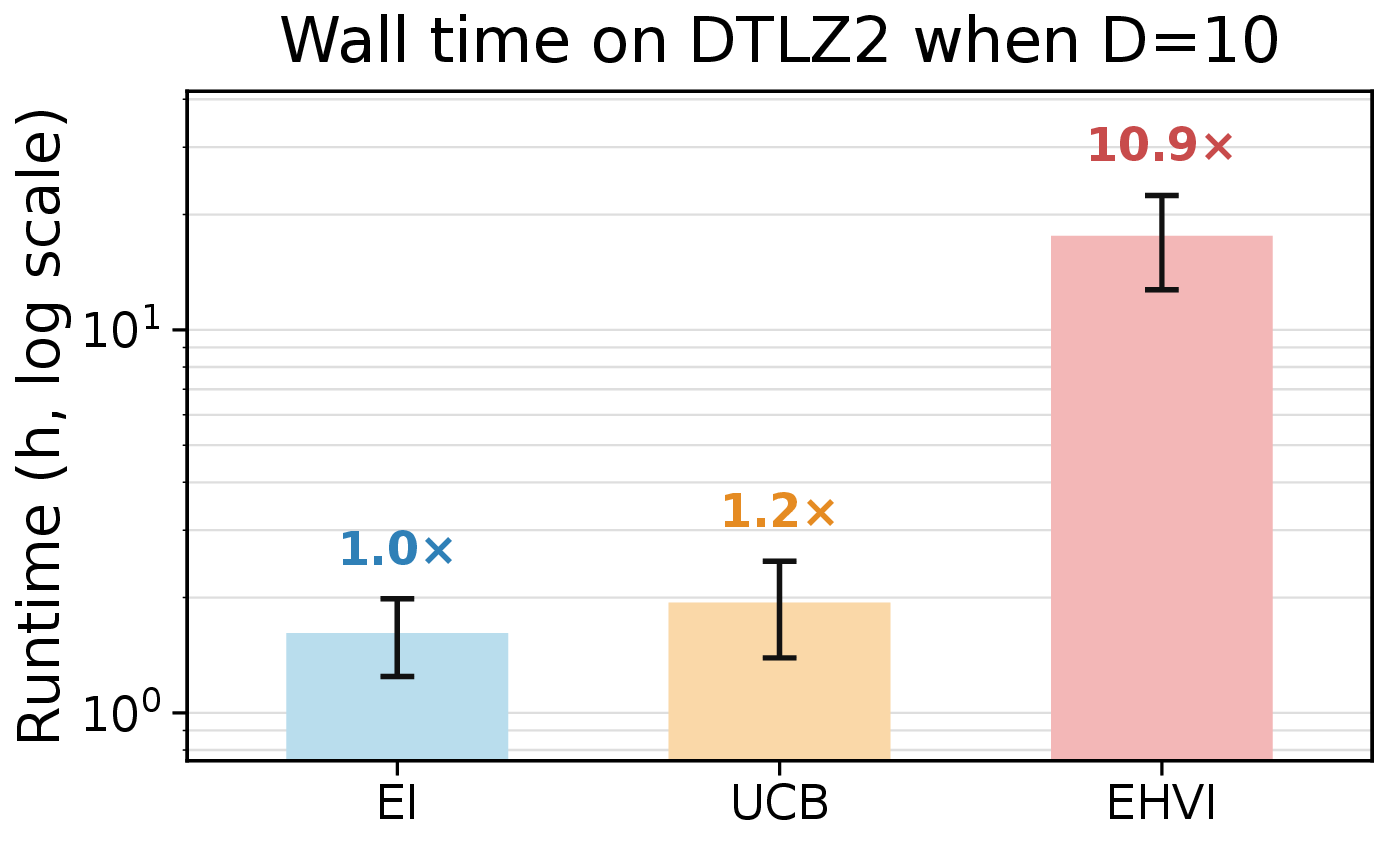}
}
\caption{Comparison of HV (Left) and Wall time (Right) of three acquisition functions (EI, UCB and EHVI) under the same experimental settings.}
\label{3obj-af}
\end{figure}

Figure \ref{af_comapre} compares EI, UCB, and EHVI on the 2-objective DTLZ2 problem with \(D=10\) and \(D=50\) under identical experimental settings. It can bee seen that EHVI achieves better optimization performance because it directly maximizes the expected improvement in dominated hypervolume and is therefore naturally aligned with Pareto-front approximation. UCB, despite its computationally convenient additive form, provides a less direct measure of multi-objective improvement and consequently performs worse than EI and EHVI in this setting. This advantage, however, does not extend to the 3-objective problem. As shown in Figure \ref{3obj-af-hvs}, EHVI improves rapidly during the early stage but subsequently stagnates, whereas EI continues to improve and achieves the highest final HV. The deterioration of EHVI is associated with the rapidly increasing complexity of hypervolume partitioning and its complexity with the number of objectives. Figure \ref{3obj-wall-time} further shows that EHVI requires approximately \(10.9\times\) the runtime of EI, whereas EI and UCB require much less time. Moreover, several EHVI runs fail because of numerical instability during surrogate-model fitting. These results indicate that, although EHVI is attractive for low-dimensional problems with few objectives, EI and UCB provides a more favorable balance among convergence performance, computational efficiency, and numerical robustness for our ViaMOBO framework for high-dimensional real-world multi-objective optimization.


\section{Conclusion}

This paper proposed ViaMOBO, a high-dimensional multi-objective Bayesian optimization framework that alleviates the curse of dimensionality by exploiting decision-variable interactions. Based on the redefined concepts of variable interaction and separability for multi-objective optimization, ViaMOBO employs a binary classifier to learn the underlying interaction structure from previously evaluated samples. By inferring interactions from the objective-value relationships between perturbed decision vectors, the proposed method avoids additional evaluations of expensive objective functions. When an additive structure is identified, the resulting variable groups are incorporated into a multi-objective additive kernel, effectively reducing the search dimensionality and facilitating acquisition-function optimization. Extensive experiments on high-dimensional synthetic benchmarks and real-world problems demonstrate that ViaMOBO can effectively exploit separable structures while achieving a favorable trade-off among solution quality, sample efficiency, and computational cost.

Nevertheless, ViaMOBO is primarily tailored to problems with separable or weakly coupled decision-variable structures, and its advantages may diminish for strongly coupled and non-separable problems. Future work will extend ViaMOBO toward a more general high-dimensional MOBO framework through more expressive interaction modeling, adaptive variable grouping, and enhanced capability for handling strongly coupled and non-separable expensive multi-objective problems.
\section*{Acknowledgment}

This research was supported by the National Natural Science Foundation of China under Grant No. 52502380 and No. 52502508, and was also supported by the Postdoctoral Fellowship Program (Grade C) under Grant GZC20250905.

\bibliography{mybib}
\bibliographystyle{IEEEtran}

\end{document}